\documentclass[manuscript]{acmart}

\newcommand{\eg}[0]{\textit{e.g., }}
\newcommand{\ie}[0]{\textit{i.e., }}

\newcommand{\revision}[1]{{\color{black} #1}}

\usepackage{comment}
\usepackage{stfloats}
\usepackage{microtype}
\usepackage[utf8]{inputenc}  
\usepackage[T1]{fontenc}     

\AtBeginDocument{%
  \providecommand\BibTeX{{%
    \normalfont B\kern-0.5em{\scshape i\kern-0.25em b}\kern-0.8em\TeX}}}

\usepackage[autostyle,german=guillemets]{csquotes}
\makeatletter
\usepackage{tabularx}
\usepackage{hyperref,quoting}
\usepackage{url}
\quotingsetup{vskip=0pt,font={itshape,raggedright},rightmargin=0pt}
\usepackage{tikz}
\usepackage{cuted}
\usepackage{capt-of}
\usepackage{subcaption}
\usepackage{fancyvrb}
\usepackage{gensymb}
\usepackage{float}
\usepackage{stfloats}
\usepackage{tikz}
\usepackage{xcolor}

\definecolor{navy}{RGB}{22,40,84}
\newcommand{\navysquare}[1]{%
    \begin{tikzpicture}[baseline=(char.base)]
        \node[draw=navy, fill=navy, circle, minimum size=1.1em, inner sep=0pt, text=white] (char) {\textbf{#1}};
    \end{tikzpicture}%
}
\definecolor{purple}{RGB}{102,102,255} 
\newcommand{\purplesquare}[1]{%
    \begin{tikzpicture}[baseline=(char.base)]
        \node[draw=purple, fill=purple, circle, minimum size=1.1em, inner sep=0pt, text=white] (char) {\textbf{#1}};
    \end{tikzpicture}%
}
\definecolor{pink}{RGB}{255,102,102} 
\newcommand{\pinksquare}[1]{%
    \begin{tikzpicture}[baseline=(char.base)]
        \node[draw=pink, fill=pink, circle, minimum size=1.1em, inner sep=0pt, text=white] (char) {\textbf{#1}};
    \end{tikzpicture}%
}
\definecolor{blue}{RGB}{115,157,211} 
\newcommand{\bluesquare}[1]{%
    \begin{tikzpicture}[baseline=(char.base)]
        \node[draw=blue, fill=blue, circle, minimum size=1.1em, inner sep=0pt, text=white] (char) {\textbf{#1}};
    \end{tikzpicture}%
}
\definecolor{darkgrey}{RGB}{130, 130, 130} 
\newcommand{\darkgreysquare}[1]{%
    \begin{tikzpicture}[baseline=(char.base)]
        \node[draw=darkgrey, fill=darkgrey, circle, minimum size=1.1em, inner sep=0pt, text=white] (char) {\textbf{#1}};
    \end{tikzpicture}%
}
\newcommand{\tool}{\texttt{U-Define}} 
\newcommand{\ours}{\tool}

\setcopyright{none}
\acmDOI{}
\acmISBN{}

\renewcommand{\footnotetextcopyrightpermission}[1]{}
\begin{document}


\title{\revision{\texttt{\tool{}}: Designing User Workflows for Hard and Soft Constraints in LLM-Based Planning}}
\thanks{This manuscript is currently under review.}
\fancyfoot{}








\author{Christine P Lee}
\authornote{Corresponding author.}
\orcid{0000-0003-0991-8072}
\affiliation{%
  \institution{University of Wisconsin--Madison}
  \country{USA}
}
\email{cplee5@cs.wisc.edu}

\author{Xinyu Jessica Wang}
\orcid{0009-0002-5519-8432}
\affiliation{%
  \institution{University of Wisconsin--Madison}
  \country{USA}
}
\email{xwang2775@wisc.edu}

\author{Aws Albarghouthi}
\orcid{0000-0002-9456-1495}
\affiliation{%
  \institution{University of Wisconsin--Madison}
  \country{USA}
}
\email{aws@cs.wisc.edu}

\author{David Porfirio}
\orcid{0000-0001-5383-3266}
\affiliation{%
  \institution{George Mason University}
  \country{USA}
}
\email{dporfiri@gmu.edu}

\author{Bilge Mutlu}
\orcid{0000-0002-9456-1495}
\affiliation{%
  \institution{University of Wisconsin--Madison}
  \country{USA}
}
\email{bilge@cs.wisc.edu}
\renewcommand{\shortauthors}{}

\begin{abstract}

\revision{LLMs are increasingly used for end-user task planning, yet their black-box nature limits users' ability to ensure reliability and control. While recent systems incorporate verification techniques, it remains unclear how users can effectively apply such rigid constraints to represent intent or adapt to real-world variability. 
For example, prior work finds that hard-only constraints are too rigid, and numeric flexibility weights confuse users.
We investigate how interaction workflows can better support users in applying constraints to guide LLM-generated plans, examining whether abstracting strictness into high-level types (\ie hard and soft) paired with distinct verification mechanisms helps users more reliably express and align intent.
We present \ours{}, a system that lets users define constraints in natural language and categorize them as either hard rules that must not be violated or soft preferences that allow flexibility. \ours{} verifies these types through complementary methods: formal model checking for hard constraints and LLM-as-judge evaluation for soft ones. 
Through a technical evaluation and user studies with general and expert participants, we find that user-defined constraint types improve perceived usefulness, performance, and satisfaction while maintaining usability. These findings provide insights for designing flexible yet reliable constraint-based workflows.}

\end{abstract}





\begin{teaserfigure}
    \includegraphics[width=\textwidth]{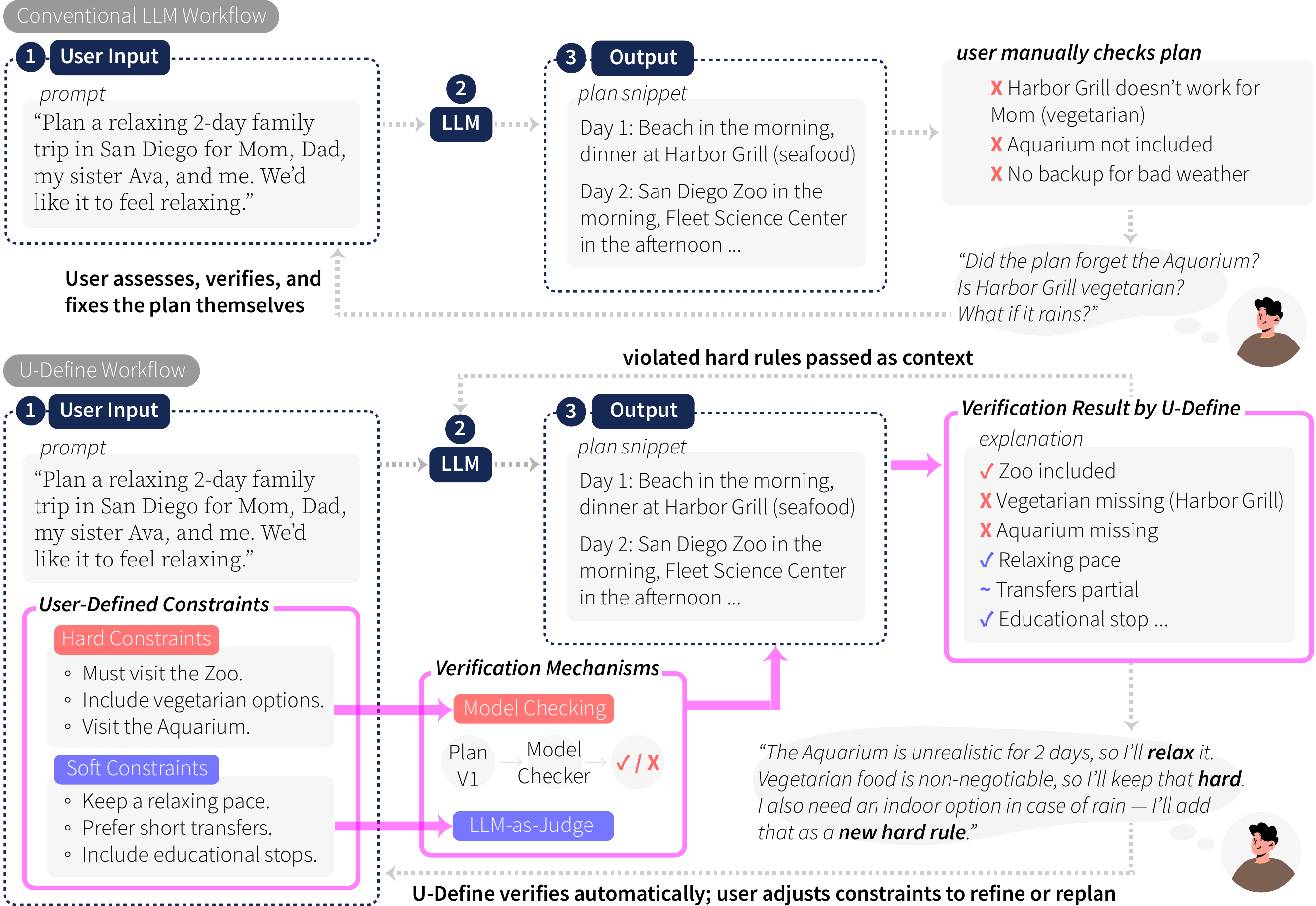}
   \vspace{-12pt}
  \caption{In a conventional workflow (top), users receive an LLM plan but must manually detect errors and missing preferences. In contrast, \texttt{U-Define} (bottom) enables users to explicitly state their own constraints in natural language, distinguishing between \textbf{hard} rules that must never be violated (\eg ``include vegetarian options'') and \textbf{soft} preferences that reflect flexibility (\eg ``keep a relaxing pace''). These are then verified by mechanisms tailored to their type: formal model checking for hard constraints and LLM-as-judge evaluation for soft ones. The system produces verification results on which constraints are met or violated, allowing users to iteratively refine, relax, or strengthen constraints. \ours{} enables users to leverage constraint types that combine reliability with flexibility.}
  \label{fig:teasor}
\end{teaserfigure}
\maketitle



\section{Introduction}

Real-world planning rarely consists of uniformly strict rules. Instead, people balance \textbf{hard constraints} (\ie rules that must never be violated) with \textbf{soft constraints} (\ie preferences that reflect flexibility and individual choice). For example, a manager creating schedules may need to enforce the rule ``a worker can only take one shift at a time'' while also considering a preference such as ``having two consecutive days off.'' Similarly, in everyday life, cooking for someone with a gluten sensitivity (a strict rule) still leaves room to choose between spaghetti, linguine, or penne (a preference). These scenarios raise a broader design question: how should user–AI interactions with black-box AI models be designed so that users can reliably express and balance their different types of constraints?

Traditional approaches to planning have largely emerged from the \textit{automated planning} domain, where the focus is on constructing action sequences that achieve a goal state while satisfying a set of constraints \cite{kress2018synthesis}. Such methods are effective in expert settings such as robotics, logistics, or manufacturing \cite{garrett2021integrated, sreedharan2019model, nau2005applications}, but they are often inaccessible to end users.
They rely on specialized planning languages, whereas everyday users need simple ways to express their intentions; they require detailed domain specifications, while users often operate with only partial or high-level expectations; and they depend on deterministic solvers, in contrast to users' desire to flexibly adjust, relax, or prioritize among rules and preferences \cite{hurnaus2010programming, porfirio2018authoring, sauer2022structure, porfirio2020transforming}. While these approaches guarantee correctness, they do not scale beyond specific domains or align with how people actually want to articulate and adapt their constraints in practice.

Recent advances in large language models (LLMs) have dramatically shifted this landscape. LLMs are attractive for planning tasks because they are accessible and flexible, and they generalize well to new contexts \cite{pallagani2024prospects, silverLLM2024, songLLMPlan2023, lu2023plug}. Yet in their off-the-shelf form, they remain unreliable and introduce challenges for users. Unlike automated planners, LLMs are known to produce inconsistent plans, fail to enforce strict requirements, and present outputs with fluent confidence that can obscure underlying errors \cite{lin2021truthfulqa, ji2023survey, kabir2023answers}. These reliability gaps are particularly detrimental in complex or high-stakes settings where violating a hard rule can have serious consequences \cite{kaddour2023challenges, rao2023evaluating}. Thus, while LLMs lower the barrier for end-user planning, their black-box nature raises significant challenges for system and interaction design: users must often shoulder the burden of manually verifying outputs, existing verification and explanation techniques for LLM-based planners remain brittle, and there are few ways for users to control or specify the constraints that matter to them.

Emerging research has begun to address these issues by combining LLMs with deterministic verification approaches such as external critics, structured pipelines, and formal verification methods \cite{kambhampati2024llms, gundawar2024robustplanningllmmoduloframework, lee2025veriplan}. For example, \citet{kambhampati2024llms} proposed the LLM-Modulo framework, which leverages LLMs with external critics, verifiers, and human input to improve planning effectiveness. In previous work, \citet{lee2025veriplan} applied formal verification techniques to assess whether LLM-generated plans satisfy user-specified constraints. These approaches provide reliable guardrails but often rely on pre-engineered templates or domain-specific specifications \cite{lee2025veriplan, guan2023leveraging, xie2024travelplanner, yang2024plug}, limiting their generality and flexibility. As these verification approaches mark progress for users' reliable LLM use, more importantly, the interaction design and user workflows for LLMs augmented with deterministic verification remain underexplored.

\revision{In response, we ask how constraint-based AI systems can better support users in expressing constraints that reflect varied intents and adapt to changing needs. Prior work highlights that users struggle with \textit{constraint flexibility}---the changing strictness of constraints often depending on evolving contexts or user preferences \cite{lee2025veriplan}. Relying solely on hard constraints is often too rigid to capture nuance, while in other situations a rule may need to be treated as a firm requirement. Some systems attempt to address this by allowing users to assign numeric weights to indicate strictness \cite{lee2025veriplan, porfirio2024polaris}, but such controls are shown to be difficult to interpret for users, as they may not know how values map to enforcement levels or which numbers to select. Hard constraints provide reliability, yet overly rigid or precise controls can overwhelm users.
We therefore focus on designing interaction paradigms and workflows that help users more effectively leverage user-defined constraints based on their needs. This motivation leads to the following research questions: (1) \textit{\textbf{Constraint expression}: How can users effectively express constraints of different strictness in LLM-based planning tasks?}; (2) \textit{\textbf{Interaction design}: How can we support users in engaging with LLMs that integrate distinct verification mechanisms for different constraint types?}; and (3) \textit{\textbf{Understanding Use}: How do different constraint types influence the quality of LLM outputs and the overall user experience?}}

\revision{Building off the advances and limitations of existing work, we propose that effective interaction design for AI systems integrating deterministic approaches requires two things: supporting users in defining their own constraints without relying on domain-specific templates, and providing workflows that support constraint flexibility and reliability.} Our system, \ours{} (\textbf{U}ser-\textbf{Define}d constraints), operationalizes this idea by introducing a workflow in which users: (1) along with their task prompt, freely input natural language constraints that are automatically translated into verifiable forms; (2) define constraints as either hard or soft; (3) see these constraints verified through mechanisms tailored to their constraint type---formal model checking for hard constraints and LLM-as-judge evaluation for soft ones; and (4) iteratively refine or adjust constraints through feedback. By giving users direct control over constraint types, \ours{} offers both the reliability of deterministic enforcement and the flexibility of LLM-based planning.

We investigate this approach through the design and prototype implementation of \ours{} expanded from \cite{lee2025veriplan} and evaluate its effectiveness through a technical evaluation, a user study with general users, and a study with domain experts. Across these studies, our findings suggest that supporting user-defined constraint types is both desired and useful. We found that users perceived hard and soft constraints as serving distinct roles in their planning tasks; the use of both improved LLM performance and usefulness without reducing usability; and hard rules were viewed as being especially important to ensure reliable plans. The contributions of this work are as follows: 
\begin{enumerate}
\item \textit{Technical contributions:} We introduce an automated pipeline that translates natural language hard constraints into formal verification artifacts using LLMs. This enables constraint enforcement without pre-engineered domain specifications.
\item \textit{System contributions:} We present \ours{}, a system that enables users to simultaneously define soft and hard constraints in their planning requests to LLMs. Depending on the constraint type, \ours{} applies tailored verification approaches: model checking for hard constraints and LLM-as-judge techniques for soft constraints.
\item \textit{Empirical contributions:} We provide evidence from technical and user evaluations that supporting user-defined constraint types improves perceived performance, usefulness, and overall satisfaction in LLM-based planning tasks.
\item \textit{Design implications:} We present design insights on how constraint flexibility can be supported in practice, how users envision workflows that integrate constraint types, and how deterministic and generative approaches can be combined to better align LLM outputs with diverse user needs.
\end{enumerate}

\section{Related Work}
In this section, we provide background on the challenges users face when using LLMs as end-user planning tools, along with existing approaches to address these issues. We then give a broad overview of automated planning and constraint-based methods, and their relevance to our goal of balancing rigor and flexibility. Finally, we review model checking and how it is used in formal verification.

\subsection{User Challenges and Approaches for LLMs in End-User Planning Tasks}

LLMs have demonstrated impressive capabilities across a broad range of tasks, from summarization and question answering to dialogue generation and reasoning \cite{gpt-4, hadi2023survey}. Their increasing deployment in real-world applications highlights their potential utility across diverse domains \cite{scanlon2023chatgpt, kim2024understanding}. However, despite their usefulness, LLMs remain imperfect, particularly when applied in high-stakes or complex real-world settings. Much of this imperfection stems from user-centered challenges, most notably, issues around reliability and transparency, that hinder user trust and broader adoption.

A key concern in the real-world deployment of LLMs is their lack of reliability. Prior work has shown that LLMs often produce inconsistent answers, even when given the same prompt repeatedly \cite{borji2023categorical, openai2023gpt4systemcard, elazar2021measuring, alkhamissi2022review}. Such inconsistency introduces uncertainty and unreliability that can undermine user confidence and satisfaction, especially in tasks requiring stable reasoning or reproducibility \cite{hendrycks2021unsolved, kim2024understanding}. Furthermore, LLMs are known to generate hallucinations, outputs that are nonfactual, nonsensical, or completely irrelevant to the prompt \cite{openai2023gpt4systemcard, ji2023survey, huang2025survey}. Such unreliability limits their effectiveness in scenarios that demand precision, consistency, or factual correctness \cite{cheong2024not, cascella2023evaluating, valmeekam2022large}, such as healthcare diagnosis \cite{kanjee2023accuracy, walker2023reliability, singhal2023large}, judicial opinions \cite{davis2023lawyer, neumeister2023lawyer}, or planning tasks based on resource availability or constraint contingencies \cite{he2025plan, lee2025veriplan, zhang2024cfbench, xie2024travelplanner}. 

Another critical challenge is the opaque, black box nature of LLMs, which significantly affects users' ability to interpret, assess, or trust the outputs \cite{barman2024beyond, liao2023ai}. LLMs typically communicate uncertainty, rationale, or system status in ways that are not easily verifiable, hindering users from identifying the source of errors, assessing the validity of responses, or determining when to intervene. This lack of transparency further exacerbates usability challenges, especially in domains where users need to make informed decisions based on model outputs \cite{kaddour2023challenges, rao2023evaluating}.

To address these limitations, a growing body of research has proposed methods and system designs to enhance LLM reliability, transparency, and alignment with user expectations. 
One line of work focuses on augmenting LLMs by utilizing more LLMs, such as using LLMs as guidelines, judges, or critics to evaluate the output \cite{mostajabdaveh2024optimization, liang2024improving, vicuna2023, wang2022self, bai2022constitutional, ayyamperumal2024current}, or design multi-agent orchestration for each agent to collaboratively play a role and enhance the output \cite{talebirad2023multi, wu2023autogen, lee2025map}. While utilizing LLMs for evaluation can be effective, there are existing concerns in using inherently unpredictable LLMs to evaluate LLMs, as the evaluation can suffer from bias based on the order, appearance, or length of the content, aspect-specific evaluation, scalability, and effectiveness in diverse contexts \cite{wang2023large, huang2024empirical, koo2023benchmarking, son2024llm, park2024offsetbias}.

Another line of work focuses on improving the technical consistency of LLMs themselves, by using different decoding strategies for reasoning \cite{wang2022self} or improving the transparency through explanations or uncertainty \cite{kuhn2023semantic}. Other work has focused on advancing the transparency of LLMs through explanations, sources, and guidelines to foster appropriate reliance \cite{kim2025fostering, jiao2024navigating}. More recent work has focused on applying constraint-based approaches to guide LLM behavior, especially in the automated planning domain \cite{yang2024plug, zhang2024cfbench, xie2024travelplanner}. 

Recently, an increasing body of work has focused on how to appropriately leverage the capabilities of LLMs in components of system designs to overcome their limitations. In this direction, some prior work explores user-in-the-loop approaches, incorporating user feedback or control to guide model behavior or flag uncertain outputs \cite{wang2024LLM, ma2024beyond, lee2025veriplan}. Other work has explored modular architectures, where LLMs are integrated as specialized components for specific sub-tasks, often combined with external verifiers or human experts to provide more structure and interoperability \cite{kambhampati2024llms, valmeekam2022large, lee2025veriplan, shankar2024validates}. For example, \citet{gundawar2024robustplanningllmmoduloframework} presents rule-based functions as a critic to evaluate LLM-generated plans, while \citet{lee2025veriplan} applies model checking techniques to verify LLM-generated plans against user constraints. Additionally, \citet{laban2024beyond} integrates LLMs with external search engines as verifiers of the LLM's outputs. 
However, existing constraint-based, logic-based, or formal verification approaches with LLMs often utilize predefined datasets or domain-specific configurations and can suffer from the lack of mechanisms for dynamically incorporating users' specific, evolving preferences and needs. As a result, there remains an open challenge in designing LLM systems that not only meet formal correctness criteria but also flexibly support the complexity and variability inherent in real-world human-AI interactions.



\subsection{Automated Planning and Constraint-Based Approaches}

\textit{Automated planning} refers to techniques for automatically generating a sequence of actions that enables an agent to achieve a specific goal from its initial state \cite{ghallab2016automated}. These techniques specifically focus on \textit{what} to do, namely defining the steps or actions that are required to reach the goal state, rather than focusing on the procedures for \textit{how} to perform each step. Existing languages and libraries (\eg \textit{Planning Domain Definition Language} (PDDL) \cite{fox2003pddl2}, the \textit{GTPyhop} planner \cite{nau2021gtpyhop}, and the extensive \textit{Unified Planning} library \cite{kapellosaiplan4eu}) support users to interact with and utilize planning algorithms for various tasks, such as robot navigation, games, and scheduling operations \cite{amer2021automated, duarte2020survey, hayamizu2021guiding}. These planning tools are intended for expert use, requiring expert knowledge such as low-level formal languages, planning domain modeling (\eg specifying predicates, actions, initial state and goal conditions), and search and optimization concepts. Recent work has focused on supporting user engagement with these planning tools through visualization \cite{DePellegrin_Petrick_2024}, user involvement in plan creation \cite{porfirio2024polaris}, and multimodal interfaces \cite{porfirio2023sketching} to name a few examples.

In automated planning and similar fields, constraint-based approaches refer to techniques where problems are modeled as a set of constraints (\ie conditions or properties that must be satisfied) rather than solved through explicit procedural steps \cite{ghallab2025acting}. In such approaches, variables represent the unknowns in a problem, and constraints, expressed as logical or mathematical statements, define the relationships or properties that must hold among them. This methodology has been widely applied across domains such as program repair (\eg synthesizing code changes to pass failing tests) \cite{le2019automated, nguyen2013semfix,xuan2016nopol,van2018static, porfirio2020transforming}, test input generation (\eg through symbolic execution) \cite{nguyen2013semfix}, scheduling and planning (\eg satisfying resource and precedence constraints) \cite{fox2003pddl2, do2003sapa}, and model checking (\eg verifying that system behaviors adhere to temporal properties) \cite{porfirio2020transforming, porfirio2018authoring}. These approaches offer a declarative, automated, and scalable method for guiding program behavior, enabling users to specify what outcomes are desired while leaving the implementation details to solvers.

A central distinction within constraint-based methods is between hard constraints and soft constraints \cite{dechter2003constraint}. \textit{Hard constraints} are strict conditions that must be met for a solution to be considered valid, as violating them renders the solution unacceptable. For instance, path conditions in symbolic execution or test cases in program repair act as hard constraints that must be fulfilled \cite{cadar2013symbolic, le2012systematic}. In scheduling, task ordering or resource availability might constitute hard rules that cannot be violated \cite{michael1995scheduling}.
In contrast, \textit{soft constraints} represent preferences or desirable outcomes that are not strictly required. Violating a soft constraint does not invalidate a solution but may make it less optimal or less aligned with user goals. Examples include minimizing code edits in program repair or preferring certain times or locations in scheduling tasks \cite{lewis2008survey, martinez2016astor}. In practice, systems often aim to satisfy as many soft constraints as possible or minimize the cost of their violation \cite{bistarelli1997semiring}.

While LLMs are not traditionally designed within a constraint-based framework, the notion of guiding behavior through structured input is loosely mirrored in their in-context learning capabilities. Prompts and examples serve as guidelines for desired behavior, shaping the output based on contextual cues \cite{gpt-4, openai_chatgpt, schulhoff2024prompt}. However, these ``constraints'' are neither guaranteed nor enforced in a systematic or deterministic way. As such, they are more akin to soft constraints, as they influence the model's behavior but adherence is inherently unpredictable and non-deterministic.

In this work, we aim to combine the strengths of constraint-based reasoning and LLM adaptability to create a more reliable yet flexible verification framework. We propose a hybrid approach that leverages the systematic and deterministic nature of formal verification for enforcing hard constraints, while also accommodating the contextual and adaptive behavior of LLMs to address soft preferences and user-defined guidelines. This integration aims to bridge the gap between rigid verification requirements and the fluid, nuanced nature of real-world user needs.

\subsection{Model Checking in Formal Verification and LTL Constraints}
Formal verification is the process of mathematically proving the correctness of a system, and \textit{model checking} is a specific technique used to carry out this task. In model checking, a system is analyzed to verify whether it satisfies a set of desired logical properties \cite{baier2008principles}.

A model checker is a software tool that performs model checking and typically involves three steps \cite{clarke2011model}: first, it takes in a system model, which is a formal representation of the system's states and the transition relation between states. Next, it takes a specification (\ie a set of logical properties, such as those expressed in temporal logic) that defines the expected behavior. Finally, it performs model checking by exhaustively exploring all possible behaviors of the model to verify whether it meets the specifications.

Linear temporal logic (LTL) is a formal language frequently used to specify temporal properties over system executions \cite{pnueli1977temporal, kress2018synthesis}. It supports a range of constraint types, including precedence constraints (\eg requiring that an action such as unlocking a door occurs before entry), eventuality (\eg ensuring that every request is eventually followed by a response), safety properties (\eg prohibiting the system from entering an error state), and invariance (\eg maintaining that a condition, such as a temperature threshold, holds at all times) to name a few. These expressive temporal operators make LTL well-suited for verifying and enforcing rule-based behaviors in time-sensitive applications. Because of their temporal expressiveness, LTL properties are particularly well-suited for and explored in domains where the timing and order of actions are critical—such as scheduling, safety protocols, and manufacturing workflows. As a result, it has been applied in domains including social robot interactions \cite{porfirio2018authoring}, assistive robotics \cite{dixon2014fridge}, and autonomous navigation \cite{liu2023grounding}.
In model checking with LTL, the goal is to verify whether all possible paths in a system, represented as sequences of states, satisfy these temporal properties.

In summary, our work builds on existing approaches to make LLMs effective end-user planning tools by integrating them with constraint-based and formal verification methods.
We address persistent challenges in the usefulness and reliability of LLMs as planning tools by allowing users to define varying levels of constraint strictness. 
Extending prior work in verification and modular design, we apply formal verification techniques alongside LLM-based evaluation methods to assess plans against user-defined constraints. Our contribution extends current integration approaches of formal verification techniques by automating the translation of natural language constraints into formal properties, overcoming the limitations of predefined templates, static datasets, or domain-specific rule sets.

\section{Improvements from \texttt{VeriPlan} \cite{lee2025veriplan}}

\revision{Our previous system, \texttt{VeriPlan}, applied formal verification techniques, specifically model checking, to ensure that LLM-generated plans adhered to user-defined constraints. These constraints were derived from a set of pre-engineered templates that users filled in to specify rules, which were then translated into Linear Temporal Logic (LTL) formulas for verification. While this approach demonstrated the potential of combining LLMs with deterministic verification, it required users to conform to rigid templates and offered limited flexibility in how constraints were expressed or controlled.

\ours{} extends and generalizes \texttt{VeriPlan} in three major ways:

\begin{itemize}
    \item \textit{Beyond template-based constraint specification.} Instead of requiring users to select from predefined templates, \texttt{U-Define} integrates LLMs to automatically translate natural language constraints into verifiable LTL properties. This removes the dependency on template structures and allows users to express constraints more freely in natural language.
    
    \item \textit{New user control mechanism through constraint types.} The prior system used a \textit{flexibility slider} to let users numerically adjust the enforcement strength of each constraint. While conceptually useful, users found it difficult to set a precise strictness value and interpret how numerical values affected constraint enforcement. In \ours{}, this mechanism is replaced with a more abstracted distinction between \textit{hard} and \textit{soft} constraint types, each associated with distinct verification methods (formal model checking and LLM-as-judge evaluation, respectively).
    
    \item \textit{Automated translation pipeline for model checking.} \ours{} incorporates an LLM-based translation layer that automates the conversion of user-specified hard constraints into PRISM-compatible verification artifacts, enabling formal verification without the need for manual encoding or domain-specific templates.
\end{itemize}

Together, these updates aim to make constraint definition more expressive, verification more automated, and user control more interpretable, aligned, and adaptive to everyday contexts. }

\begin{figure*}[!b]
  \includegraphics[width=\textwidth]{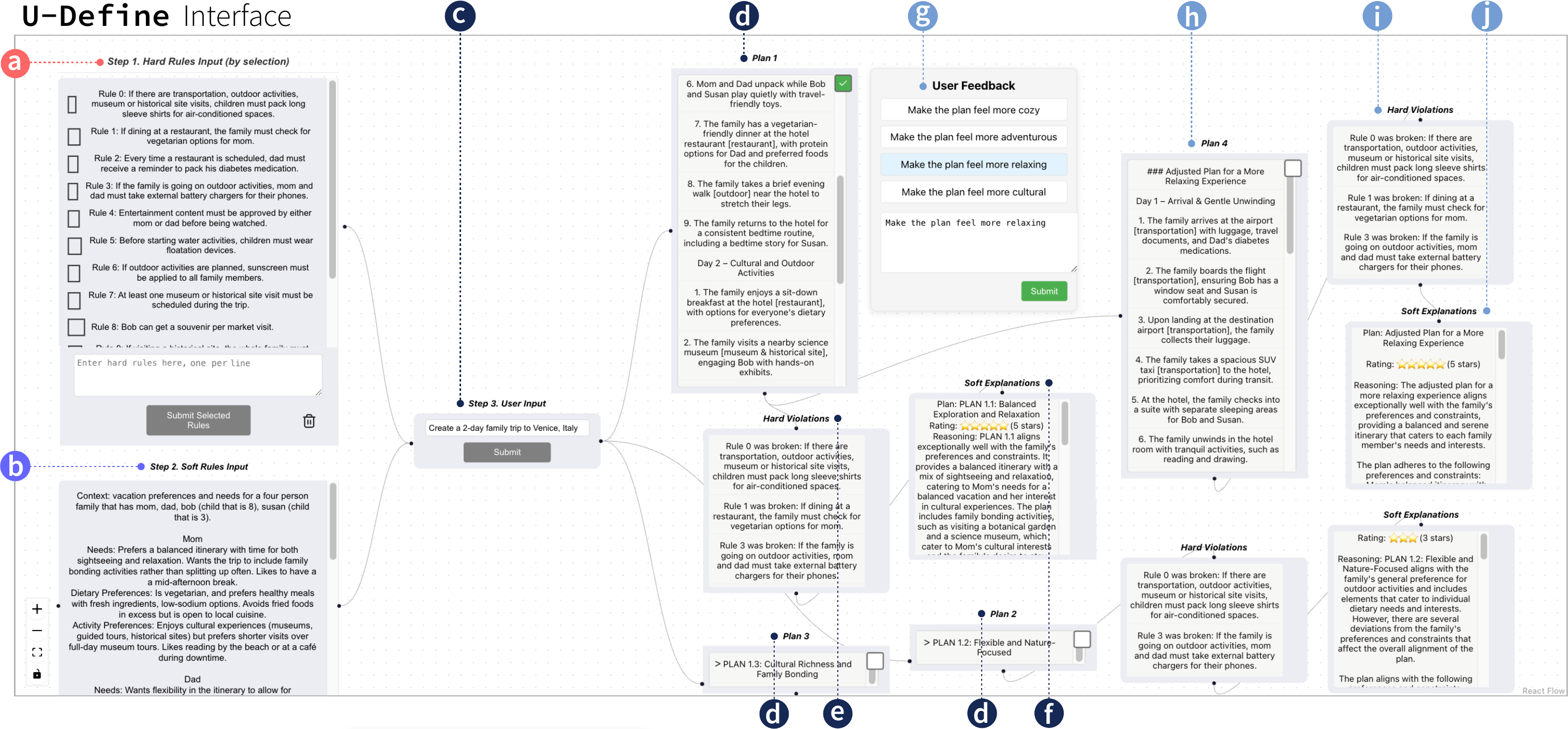}
   \vspace{-12pt}
  \caption{\textit{Front-end interface of \ours{} ---} The user begins by defining soft or hard constraints for their planning task into \ours{}. The user then inputs their request, and in response, an initial set of plans is generated. Each plan is then evaluated by distinct verification mechanisms on its adherence to the user-defined soft and hard constraints. The verification results and explanations are presented to the user. The user can select the plan they like most and provide feedback on the prompt and constraints to align the details of the plan to their needs. 
}
  \label{fig:frontend}
\end{figure*}

\section{\ours{} System Design} \label{technicalApp}

In this section, we present the system design and implementation of \ours{}. Below, we first introduce a real-world scenario to illustrate how the system operates. Building on this scenario, we describe how \ours{} structures the user's planning process into three stages: (1) the definition stage, (2) the verification stage, and (3) the feedback stage. Each stage is supported by different technical components, including the Rule Translator, LLM Planner, Model Checker, LLM-as-Judge, and Feedback Loop.


\subsection{Family Vacation Planning Scenario}\label{scenario:vacation}
To demonstrate the system in action, we present a scenario in which a user employs \ours{} to plan a family vacation. The goal is to generate a plan that supports diverse preferences and satisfies specific constraints from all family members. The scenario is illustrated through the front-end interface in Figure~\ref{fig:frontend}.

\begin{quote}
    You are a parent in a family of four---mom, dad, Bob, and Susan. Your family is planning a one-week trip to Venice, Italy, and you are in charge of the planning. You aim to create an itinerary that reflects everyone's preferences and needs. For example, preferences may include seeking adventurous outdoor activities, while needs may involve ensuring children wear flotation devices during water activities (\eg swimming pools or lakes) or taking diabetes medication after meals. 
\end{quote}
We use this scenario throughout the section to illustrate how \ours{} assists users in generating vacation plans that incorporate both hard and soft constraints. 

\subsection{Definition Stage}

In the definition stage, the user begins by inputting their constraints in natural language, which can be specified as either soft or hard types. Hard constraints undergo preprocessing before entering the verification stage. Specifically, they are passed through a translation process that converts them into LTL properties and the PRISM format required for model checking. PRISM is a widely used probabilistic model checker that supports verification of systems against temporal logic specifications \cite{KNP11}. This translation is handled by the \textit{Rule Translator}, which automatically converts natural language constraints into the necessary formal representations.

To ensure correctness, the Rule Translator then converts the final translation back into natural language and displays it to the user for validation. If the user accepts the translation, it is retained; if not, the user may delete the rule and re-enter the constraint for a new translation. Once all hard rules are finalized, all hard and soft constraints are passed to the verification stage.

\begin{figure*}[!t]
  \includegraphics[width=\textwidth]{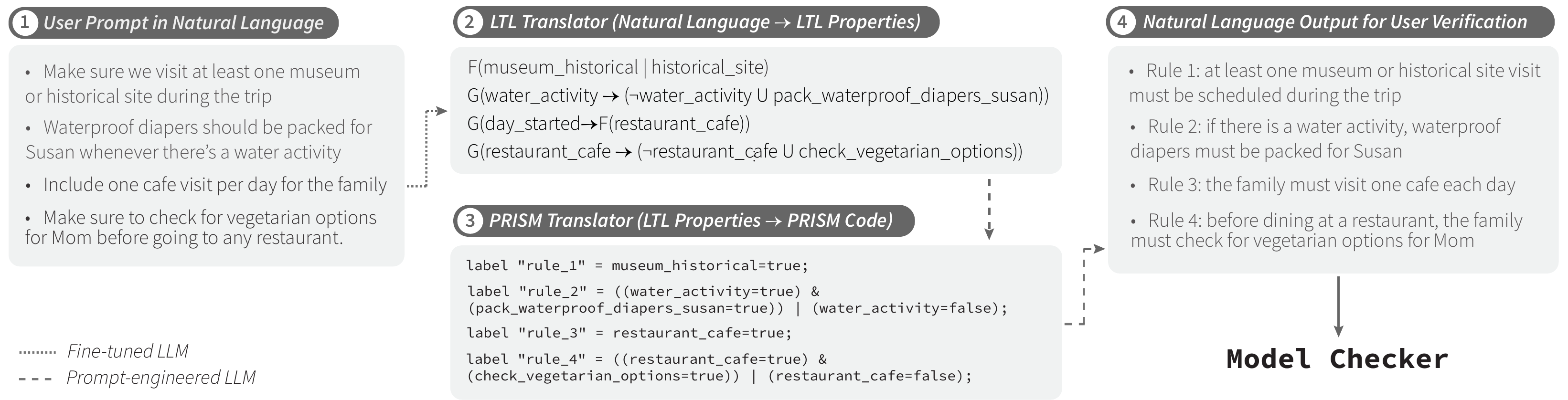}
   \vspace{-12pt}
  \caption{\textit{\ours{} Rule Translator ---} Pipeline of the rule translator described in \S\ref{finetune}. 
  The Rule Translator is an LLM-based agent fine-tuned on a natural language to LTL dataset that translates user-defined hard constraints into LTL properties for model checking. These properties are encoded in PRISM language for interpretation by the model checker, then converted back into natural language for user validation.}
  \label{fig:translator}
\end{figure*}

\subsubsection{User Workflow}\label{translator2}

At the beginning of the interaction, users input their preferences and constraints (depicted in Figure \ref{fig:frontend}). These constraints are written in natural language and can be specified as either hard or soft through the interface.
This design decision was motivated by findings that users both struggle to assign precise numeric weights yet readily express rigidity (must vs. should), and they generally prefer categorical selection or ordering over fine-grained tuning \cite{lee2025veriplan, porfirio2025uncertainty, kuhlman2019evaluating, robertson2023expressiveness}.
Once the user finishes defining their constraints, they appear in the interface as either hard (step \pinksquare{a}) or soft (step \purplesquare{b}).

\subsubsection{Technical Details of the Rule Translator} \label{finetune}
The specific pipeline of the Rule Translator is shown in Figure \ref{fig:translator}. The Rule Translator consists of two components: (1) the LTL translator and (2) the PRISM translator.

The goal of the LTL translator is to convert natural language constraints into LTL properties that are required for model checking. The translation process begins with the Rule Translator accepting the user's hard constraints in natural language (step \darkgreysquare{1} depicted in Figure \ref{fig:translator}). These constraints are then translated into LTL properties using a fine-tuned LLM agent (step \darkgreysquare{2}).


The LTL Translator translates natural language constraints into formal LTL expressions using modal operators. For example, the global operator ($G$) specifies conditions that must hold in every state; the future operator ($F$) checks for events that must eventually occur; the until operator ($U$) ensures that an event $\phi$ holds until another event $\psi$ occurs, and that $\psi$ must indeed happen; and the negation operator ($\neg$) indicates that a particular condition must not be true at a specific point or over a given temporal duration.

Once translated, the LTL properties are passed to the PRISM translator (step \darkgreysquare{3}). The goal of this component is to convert the LTL rules into a format that can be parsed by the PRISM model checker \cite{kwiatkowska2011prism}, described in detail in \S\ref{sec:model_checking}. The PRISM translator uses a prompt-engineered LLM-based agent that translates state representations and temporal conditions into the PRISM modeling language. We implemented the PRISM translator using a prompt-engineered LLM agent rather than an algorithmic approach because it integrates cleanly with our LLM pipeline and demonstrated feasibility during system design (see \S\ref{sec:techEval}).

Finally, the LTL and PRISM translations are converted back into natural language for user review and verification (step \darkgreysquare{4}). The user is presented with the translated rules and can verify whether they align with their intended constraints. If a rule is incorrect or misaligned with user intent, it can be deleted through the interface and re-entered for translation. Once the user confirms the correctness of all rules, the finalized set of hard and soft constraints is presented in the interface and passed to the verification stage.

\subsection{Verification Stage}

In the verification stage, \ours{} checks whether a generated plan adheres to the user-defined hard and soft constraints by applying different verification methods for each type. For hard constraints, \ours{} uses a formal verification technique---model checking--- to evaluate each plan and return a list of violated constraints to the user.

For soft constraints, a separate LLM agent assesses how well each plan aligns with the user's preferences. The agent provides a one-to-five-star rating to indicate the degree of alignment, along with an explanation that highlights any violations or reasoning behind the rating.


\subsubsection{User Workflow}

Once all constraints are defined, the user inputs their planning request, \textit{``help me create a two-day family trip to Venice, Italy''} (step \navysquare{c}). To support input and review, \ours{} adopts a mind-map (node–link) layout so users can externalize, group, and relate constraints while keeping an at-a-glance overview of the space. Prior HCI work shows that concept/mind mapping supports organization and integration of ideas in interactive tasks, and that digital mind-map tools aid workflow and collaborative ideation \cite{wang2019minddot, dunne2012graphtrail}.

In response, an LLM agent, which we refer as the \textit{LLM Planner}, generates three candidate plans from a prompt that instructs it to respect the constraints (step \navysquare{d}).
These plans are then passed on to distinct verification mechanisms, to check constraint adherence: \textit{model checking} for hard constraints and an \textit{LLM-as-Judge} for soft constraints.

After verification, \ours{} orders the plans by the number of hard-constraint violations, prioritizing those with fewer violations for interface presentation. The number of hard constraint violations was used as the sole ranking metric for interface presentation, as fewer violations indicate closer alignment with a fully valid plan. 
The ranked plans are displayed along with verification outputs: the list of hard-rule violations (step \navysquare{e}) and the LLM-judge ratings of soft constraints with accompanying explanations (step \navysquare{f}).

\begin{figure*}[!t]
  \includegraphics[width=\textwidth]{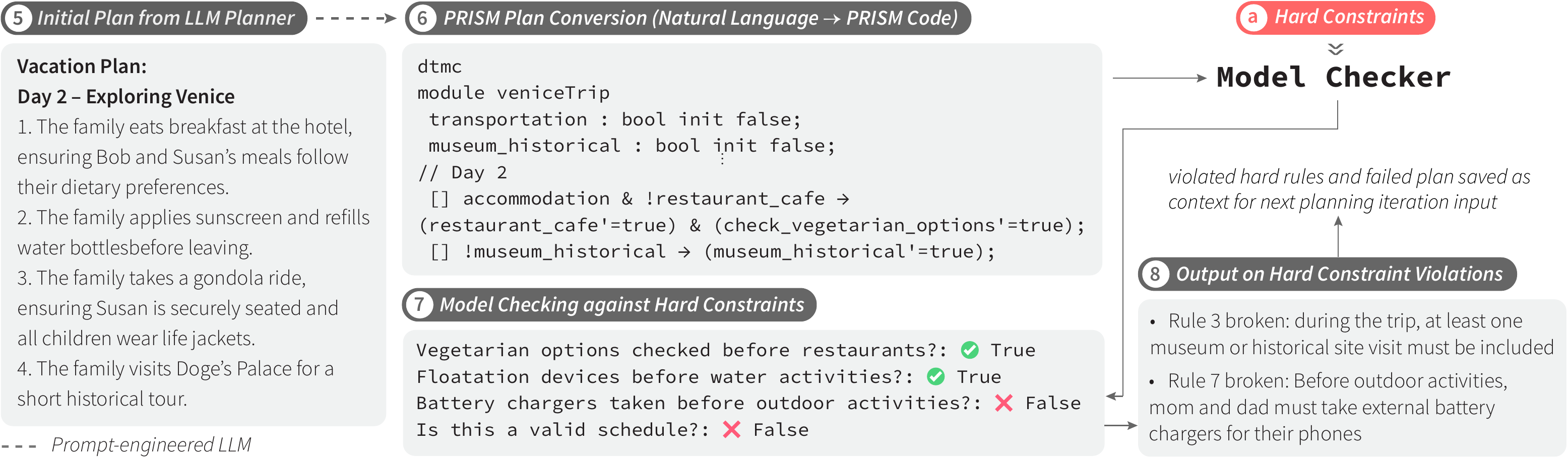}
   \vspace{-12pt}
  \caption{\textit{\ours{} Model Checker ---} Pipeline of the model checker described in \S\ref{sec:model_checking}. The model checker takes in the LLM-generated plan in PRISM language and the user-defined hard constraints from the Definition stage, and then evaluates the plan against these rules. The verification result (\ie validity of the generated plan), along with any violated rules, is presented to the user through the interface.
}
  \label{fig:modelChecker}
\end{figure*}





\subsubsection{Technical Details of the Model Checker} \label{sec:model_checking}

To verify a plan's adherence to hard constraints, \ours{} applies a formal verification technique, model checking. For verification, we use the PRISM Model Checker \cite{kwiatkowska2011prism} and Stormpy, a Python API for Storm \cite{hensel2022probabilistic}, which enables model checking within a Python environment.

Performing model checking at this stage requires two components: the hard constraints and the generated plan. The hard constraints are passed from the definition stage, ready for verification. The generated plan, however, must be converted into a representation interpretable by the model checker. Thus, the plan generated based on the user's request (step \darkgreysquare{5} in Figure \ref{fig:modelChecker}) is first translated into the PRISM modeling language using a prompt-engineered LLM agent (step \darkgreysquare{6}). We refer to this process as \textit{PRISM plan conversion}. The LLM agent translates the plan's events, state representations, and temporal conditions expressed in natural language into PRISM format.

At this stage, the model checker has access to (a) the set of hard constraints and (b) the LLM-generated plan in PRISM format. Using these components, it evaluates whether the plan satisfies the specified rules by inspecting each state to detect violations (step \darkgreysquare{7}). If any violations are found, the plan is marked as invalid and the violated rules are recorded. The model checker then outputs the verification result (\ie validity of the generated plan) along with the list of violated rules (step \darkgreysquare{8}). For hard constraints, the violated rules and failed plan are saved as context and entered as part of the prompt for the next planning iteration.

\subsubsection{Technical Details of the LLM-as-Judge}
The soft constraints are stored as rules and incorporated into a prompt for an LLM agent, which we call the \textit{LLM-as-Judge}. This agent is tasked with evaluating the generated plan against the provided rules. The LLM-as-Judge is prompt-engineered with clear instructions and example inputs and outputs to guide its evaluation. Its output consists of a five-star rating reflecting how well the plan adheres to the soft constraints, with higher ratings indicating stronger alignment, along with a descriptive explanation justifying the score.

\subsection{Feedback Stage}
Once users review the generated plans and the corresponding verification outputs, they select the plan they prefer. After selection, they can engage in a \textit{feedback loop} to iteratively refine their constraints and plans.

\subsubsection{User Workflow}
During the feedback loop, users can refine or redefine both soft and hard constraints. For soft constraints, they provide input through the feedback box displayed next to the generated plans, which also offers suggestions for soft constraint feedback (step \bluesquare{g}). For hard constraints, users can adjust them through the initial input box (step \pinksquare{a}).

\ours{} incorporates this feedback along with previous verification results to generate a revised plan (step \bluesquare{h}), re-evaluates it in the verification stage, and presents updated hard constraint violations (step \bluesquare{i}) and soft constraint explanations (step \bluesquare{j}). This process can be repeated until the user is satisfied, or applied to alternative plans from the initial set.

\subsubsection{Technical Details of the Feedback Loop}
Based on user input, the LLM Planner regenerates a plan that integrates the previous plan, verification history, and updated constraints.

Soft constraint feedback is converted into a prompt passed to the LLM-as-Judge for verification, while hard constraints are re-translated through the definition stage. The updated constraints, together with the newly generated plan, are then re-verified. \ours{} presents the updated verification results to the user, completing one cycle of the feedback loop.

\subsection{Implementation Details}
All LLM agents in our implementation are powered by GPT-4 \cite{gpt-4}, except for the single fine-tuned agent used in the Rule Translator. 
The Rule Translator used an LLM-based agent fine-tuned on the LLaMA 7B model \cite{touvron2023llama}, trained on a natural language to LTL dataset \cite{cRick_Dataset}. 
Fine-tuning was performed using QLoRA techniques \cite{dettmers2023qlora} on a Tesla T4 GPU with a standard 80/20 training/validation split. While training was planned for two epochs over 240,000 examples, we applied early stopping after approximately 1,500 steps (0.13 epochs), as the model showed early convergence. The loss curve stabilized around 0.75 with minimal fluctuation, indicating that the model had effectively learned the mapping from natural language to LTL.
At inference time, the fine-tuned LLM is further guided by a few-shot prompt containing examples of human-written LTL translations to improve output reliability.
The front end is built using React \cite{react}. Prompts for each LLM agent and the complete source code are available in the supplementary materials.\footnote{The supplementary materials can be found at \url{https://osf.io/7sgn3/overview?view_only=db6e205a5d034ec9991c25c1b316d0b9}}

\begin{table*}[!tb]
    \caption{\textit{Component-Level Evaluation ---}
    Accuracy and error analysis of the translators used in \ours{}. Details in \S\ref{sec:techEval}.}
    \label{techeval}
    \small
    \renewcommand{\arraystretch}{1.3}
    \begin{tabular}{p{0.18\textwidth}p{0.4\textwidth}p{0.07\textwidth}p{0.25\textwidth}}
         \toprule
  \textbf{Translation} & \textbf{Trained Info}& \textbf{Accuracy}& \textbf{Error Description} \\
         \midrule
 LTL Translation & LLaMA 7B \cite{touvron2023llama} fine-tuned on NL2LTL dataset \cite{cRick_Dataset} & 83.12\% & Predicate errors, operator errors \\
\hline
PRISM Plan Conversion & GPT-4 \cite{gpt-4} prompt-engineered with manual examples & 97.4\% & Predicate errors \\
          \bottomrule
    \end{tabular}
\end{table*}

\section{Component-Level Evaluation of \ours{}} \label{sec:techEval}

In this section, we present the technical evaluation of components in \ours{}.
Our goal was to assess the performance of the two LLM-based translators central to automating the model checking process: the LTL translation in the Rule Translator and the PRISM plan conversion. The LTL translator was evaluated for its accuracy in converting natural language constraints into LTL properties, while the PRISM translator was assessed based on how well it performed the PRISM plan conversion. 
Note that there is an additional step to PRISM translation done by the Rule Translator---the conversion of LTL properties produced by the LTL translator to a syntactic form that is recognizable by the PRISM model checker. This additional step is a purely syntactic conversion and is thus well-suited for LLMs, whereas LTL translation and PRISM plan conversion are much more complicated, requiring the interpretation of user intent. Thus, our evaluation focuses solely on LTL translation and PRISM plan conversion.


The purpose of our performance evaluation is to assess whether the LLM-produced translations match what a human would be reasonably expected to produce. We thereby compared the output of each translator to a human-produced counterpart created by the first author and verified for coherence by a senior author. Each human-produced counterpart reflects the authors' interpretation of the intended LTL rules and corresponding PRISM plans, and while a useful metric of human-likeness, should not necessarily be interpreted as representing ground-truth correctness. For PRISM translation, plans were treated as checklists, and comparison focused on the presence and ordering of expected steps. All ground truth translations, along with outputs from the Rule Translator and PRISM Translator, are included in the supplementary materials.

The evaluation consisted of two components. First, we calculated the Levenshtein distance between each rule or plan state and its corresponding ground truth. Levenshtein distance (or edit distance) quantifies string similarity based on the minimum number of edit operations (\ie insertions, deletions, or substitutions) required to transform one string into another \cite{sankoff1983time, yujian2007normalized}.

Second, to gain a deeper understanding of translation errors, the first author conducted qualitative coding on the outputs. Errors were categorized into three types: operator errors, such as incorrect translation of temporal operators (\eg $G$, $F$, $U$); predicate errors, involving misinterpretations of truth values or state attributes; and symbol errors, such as the mistranslation of logical symbols (\eg $\land$, $\lor$).

The Rule Translator was evaluated on 36 rules in total, with 12 rules randomly selected from each of the three user study scenarios based on the hard constraints created by participants. The PRISM Translator was evaluated using three plans, randomly selected from one participant across each scenario.
The final accuracy was derived from the average performance across these iterations, as shown in Table \ref{techeval}. We discuss the results in more detail below.

\subsection{LTL Translation Performance}
The average Levenshtein similarity across all hard constraints translated by the Rule Translator was 83.12\%.
Our qualitative analysis revealed that most errors occurred in predicate translation. In four cases, the LLM added unnecessary information to predicates. For example, for the constraint \textit{``Before any transportation, outdoor activities, or visits to museums or historical sites, children must pack long sleeve shirts for air-conditioned spaces,''} the correct predicate is \texttt{pack\_long\_sleeve\_shirts}, but the LLM generated \texttt{pack\_long\_sleeve\_shirts\_for\_air\_conditioned\_spaces}. In 13 instances, the LLM produced incorrect tokens for predicates, such as translating \textit{``Before outdoor activities, both mom and dad must take external battery chargers for their phones.''} as \texttt{mom\_external\_battery\_charger} instead of the correct \texttt{mom\_takes\_battery\_charger}.


The Levenshtein similarity score in Table \ref{techeval}, produced by these 13 instances, reflects what participants experienced in our user study. However, if we assume that such token-level mismatches can be easily corrected, such as by using LLMs' general strength in refining minor wording errors, the adjusted similarity score would increase to 94.32\%. We consider this a reasonable assumption for future iterations of \ours{}, as these mismatches involve minor phrasing differences that can be resolved by referencing available context. One possible future iteration may involve extracting the plan tokens that are created during the PRISM plan conversion. We can then refer to these tokens as the ground-truth set of all possible tokens, and directly map LLM-produced LTL predicates to tokens within this ground-truth set in order to improve predicate accuracy.
Additional errors were found in symbol usage, where four translations omitted logical operators such as $\land$ or $\lor$. For instance, the constraint \textit{``At least one snack must be given to the cook during the cooking process''} should be translated as \texttt{F(cooking\_active \& snack\_given)}, but the LLM generated only \texttt{F(snack\_given)}. These errors typically occurred in constraints involving combinations of simple logical conditions.
Finally, no errors were found in the translation of temporal operators.

\subsection{PRISM Plan Conversion Performance}

The average similarity across all planning steps for each scenario was 97.4\%, with 97.79\% for the vacation scenario (21 steps in the plan), 98.28\% for the recipe scenario (18 steps), and 96.11\% for the wedding planning scenario (16 steps). Our qualitative analysis revealed that all discrepancies occurred at the predicate level. In three cases, the translator either missed or incorrectly translated a predicate. For example, the step \textit{``Dice tomatoes and basil for bruschetta topping''} should be translated as \texttt{(chopping\_cutting' = true);}, but the LLM translated it as \texttt{(prepping\_mixing' = true);}. No errors were found in the translation of symbols or logical operators.

\section{User Study 1: Everyday Users}\label{sec:userStudy}
To first understand how \ours{} performs in everyday planning contexts, we conducted a study with general users to evaluate its usability, perceived value, and support for constraint-based planning.

\subsection{Method}\label{sec:userStudy_method}
\subsubsection{Scenarios}
We designed three end-user planning scenarios to use in the evaluation of \ours. One of these scenarios is the ``vacation planning'' scenario described in \S\ref{scenario:vacation}. Below, we describe the other two scenarios. 

\paragraph{Wedding Planing}

The user needs to plan a wedding for a couple, Sarah and James, who are getting married in a month. The wedding will be held over a weekend, including a rehearsal dinner, ceremony, and reception. The plan should include organizing the schedule, activities, and arrangements.
Example preferences may include the bride preferring a traditional ceremony, while the groom wants a shorter event with modern touches. Examples of stricter requirements may involve the bride's family traveling from abroad and needing time to rest before the ceremony, or guests including young children who need a quiet space and entertainment options, as well as elderly relatives who require accessible seating and transportation.

\paragraph{Multiple Recipe Planning}
The user needs to plan a dinner party taking place on Wednesday at 6:00 PM with multiple guests. The dinner involves preparing several dishes to accommodate different dietary needs and preferences.
The plan should include organizing the cooking schedule, ingredient preparation, and timing for serving. Example preferences may include making spaghetti and meatballs as the main dish and cheesecake for dessert, with both vegetarian and gluten-free versions. Examples of stricter constraints may include limited oven space, guests arriving early who expect appetizers, or coordinating cooking steps to ensure everything is ready and warm by 6:00 PM.

\subsubsection{Study Design}
We conducted an ablation study using a within-subjects design, where different ablation conditions served as the within-subjects variable.

In Condition 1, participants engaged with \tool{}, which included all hard and soft constraints. Condition 2 removed the hard constraints, including only the soft constraints. Condition 3 removed the soft constraints, including only the hard constraints. In Condition 4, all soft and hard constraints were removed. For consistency, we denote these conditions with \textit{C1 (\ours{})}, \textit{C2 (Soft Only)}, \textit{C3 (Hard Only)}, \textit{C4 (None)} in the remainder of the paper.

During the study, participants were randomly assigned to one of the three scenarios. In each scenario, participants engaged in all four conditions in a randomized order. After each condition, participants completed the quantitative scales and semi-structured interviews. We also categorized participants' number of iterations to fix constraints when engaging with each condition, as their usage pattern data. The entire study lasted approximately 1.5 hours, and participants were compensated \$15.00 per hour. Questionnaires used during the study can be found in the supplementary materials.

\subsubsection{Measures}
To evaluate the participants' experiences with the system, we employed the Usefulness, Satisfaction, and Ease (USE) scale \cite{article} to measure two key dimensions: usefulness (Cronbach's $\alpha = 0.90$) and satisfaction (Cronbach's $\alpha = 0.93$). We also used the performance questionnaire from the fairness, accountability, transparency, and explainability (FATE) scale developed by \citet{shin2021effects} to measure participants' perceived quality of the LLM's output (Cronbach's $\alpha = 0.88$). Finally, we measured the usability of the system using the System Usability Scale (SUS) scale \cite{brooke1996sus} (Cronbach's $\alpha = 0.87$). Usability scores were calculated by adjusting for item polarity (positive or negative) and applying the standard SUS scoring procedure. All scales were placed on a seven-point Likert scale. Thus, SUS was scored on a range from 0 to 150.

\subsubsection{Participants}
A total of 12 participants were recruited for the user study. Eligibility criteria included being located in the United States, fluency in English, and a minimum age of 18. All participants were recruited through university mailing lists.
Though constrained by a relatively small sample size, the within-subjects study design provides sufficient statistical power to detect significant effects \cite{bellemare2014statistical}.


Participants age ranged from 19--42 ($M = 28.75$, $SD = 8$). 41.6\% of the participants identified as female and 58.3\% as male. 58.3\% of our participants identified as White, 25\% as Asian, and 16.6\% as Black. We refer to participants as P1--P12, using the notation P\textit{i} to indicate participants, where \textit{i} indicates participant ID number.

\subsubsection{Analysis}\label{sec:analysis}

For the quantitative and usage pattern data, we conducted a Dunnett's test to compare the means of the ablation groups (C2, C3, C4) to the mean of the full system (C1). Dunnett's test compares the mean of several experimental conditions to a control condition, in which for our study, the full \ours{} system (C1) is considered to be the control. The test was performed with an alpha level of 0.05.

For qualitative data, we conducted a Thematic Analysis (TA) on the interviews. The coding of the responses was conducted by deriving representative themes from transcriptions~\cite{clarke2014thematic, McDonald19}. During open coding, the first author coded for significant concepts in the data. Concepts were then categorized into clusters, further being grouped into themes. These themes were iteratively discussed between the whole research team, recategorizing the groups and revising the themes upon disagreement until a consensus was reached.

\begin{figure*}[!tb]
  \includegraphics[width=\textwidth]{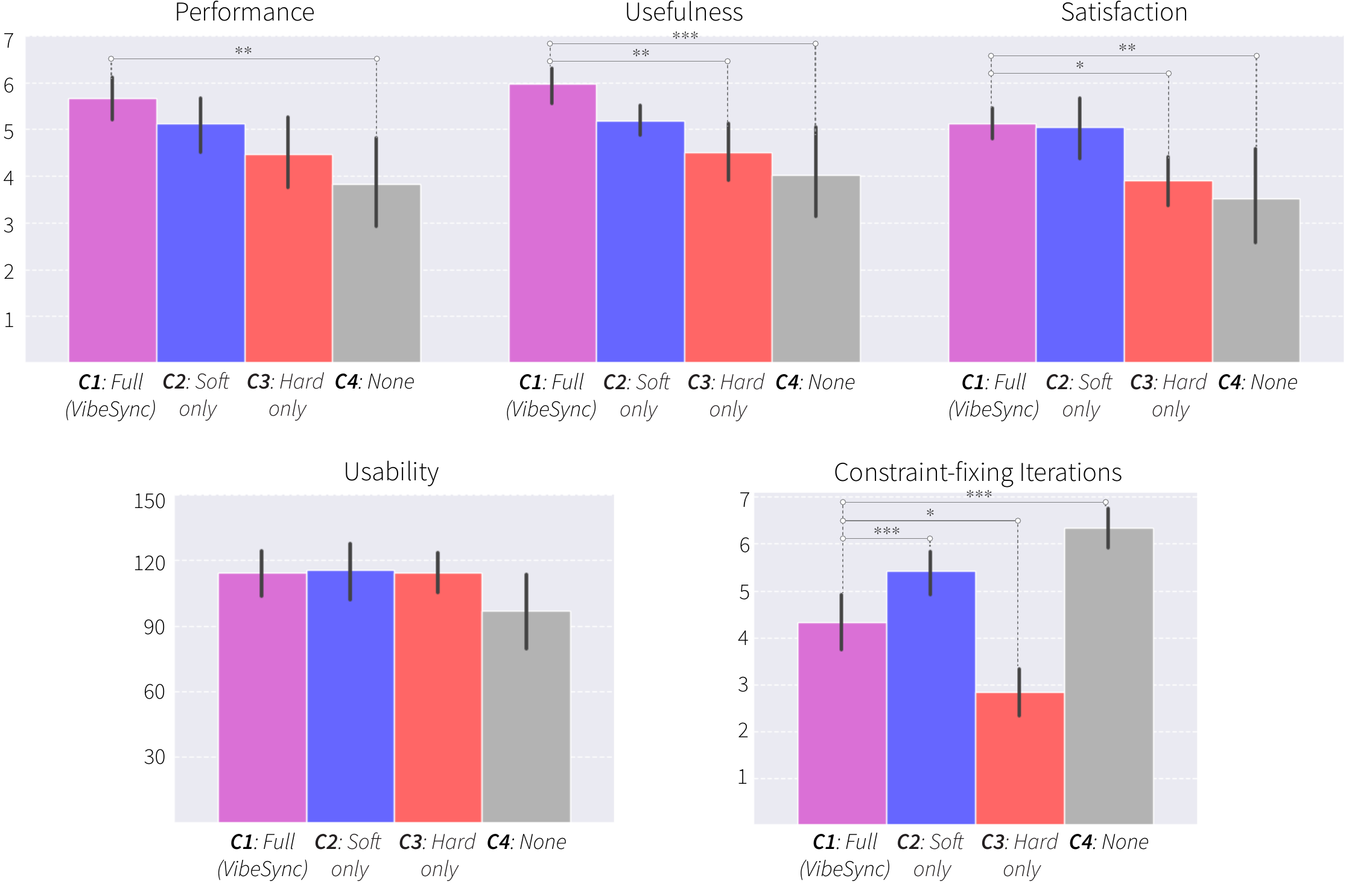}
   \vspace{-12pt}
  \caption{\textit{Quantitative Data from User Study ---} Bar graphs on participants' perceived performance of LLM, usefulness, satisfaction, usability scores, and constraint-fixing iterations across different conditions. Horizontal lines indicate significant pairwise comparisons with \revision{Dunnett's test} ($p < 0.05^{\ast}$, $p < 0.01^{\ast\ast}$, $p < 0.001^{\ast\ast\ast}$). Vertical lines in each bar graph indicate standard error. 
}
  \label{fig:userStudyQuant}
\end{figure*}

\begin{figure*}[!th]
  \includegraphics[width=\textwidth]{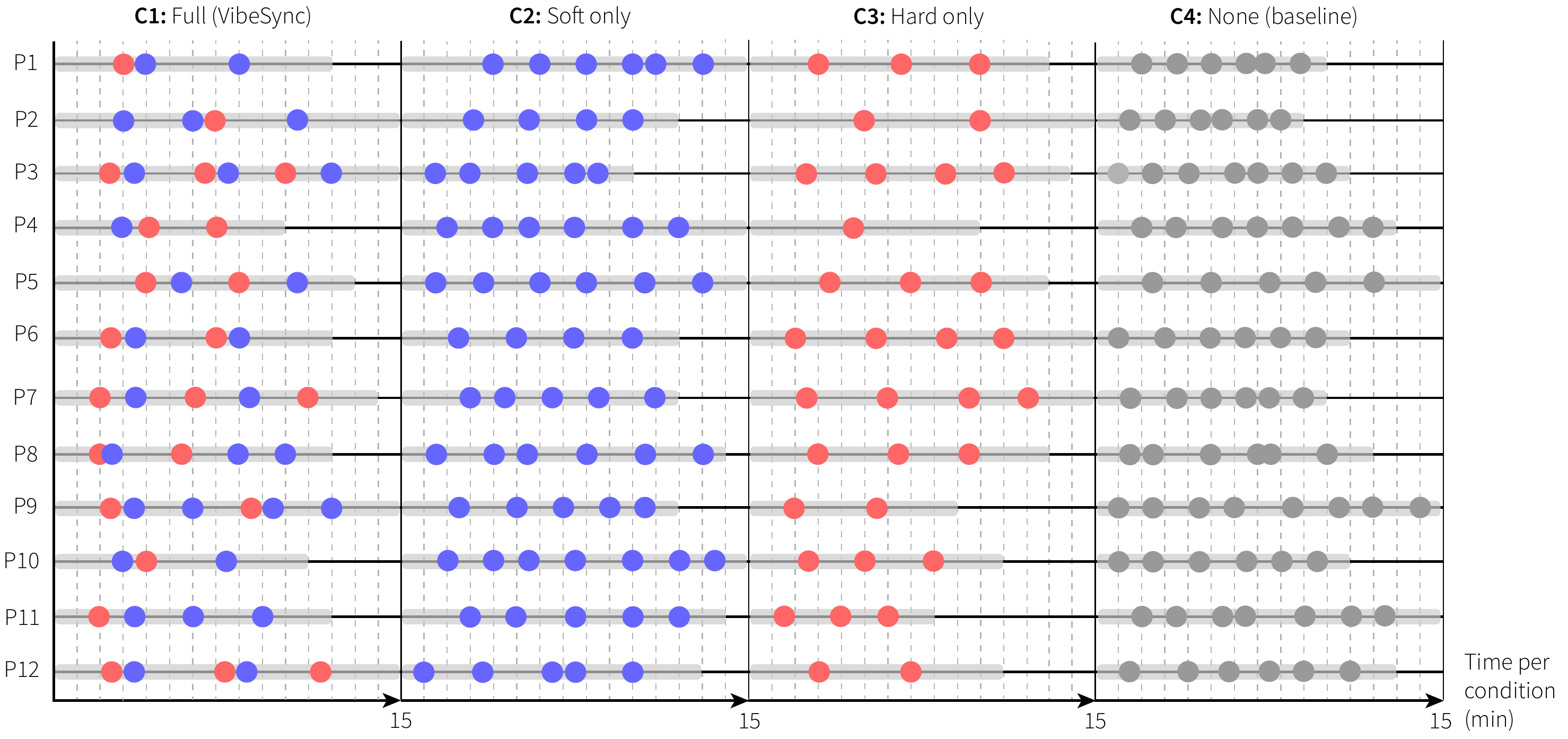}
   \vspace{-12pt}
  \caption{\textit{User interaction patterns across study conditions ---} Each dot marks when a participant added or modified a constraint during a 15-minute session. Grey bars show engagement time; conditions vary by constraint support: C1 (\texttt{U-Define}), C2 (soft only), C3 (hard only), C4 (none).
}
  \label{fig:userStudyQuant2}
\end{figure*}

\subsection{Findings}
Our analysis aimed to understand the impact of enabling users to simultaneously express and verify soft and hard constraints in LLMs for end-user planning, with a focus on effectiveness and user experience. Quantitative results are presented in Figure~\ref{fig:userStudyQuant}. Objective measures of participants' usage patterns during each condition are presented in Figure~\ref{fig:userStudyQuant2}. Derived from our quantitative, qualitative, and objective measures, we present three main findings: (1) perceived usefulness and distinct roles of constraints in \ours{}; (2) perceptions and pitfalls of using only hard constraints; and (3) adaptability support and explanation gaps in soft constraints.

\subsubsection{Perceived Usefulness and Distinct Roles of Constraints in \ours{}}\label{sec:qual1}

\paragraph{Quantitative \& Usage Pattern Results}

The Dunnett's test revealed that \textit{C1 (\ours{})} had significantly higher perceived performance $(p=.0037)$, usefulness $(p=.0003)$, and satisfaction $(p=.0052)$ compared to \textit{C4 (None)}. Participants in C1 also required significantly fewer constraint-fixing iterations than in C4 $(p<.0001)$, indicating more efficient task completion.

Our qualitative findings below suggest that these improvements stem from the complementary roles of hard and soft constraints, which together helped participants complete tasks more fully and with fewer adjustments.

\paragraph{Qualitative Findings}

All participants consistently valued the clear delineation between soft and hard constraints, noting that both played distinct but complementary roles in supporting constraint flexibility. Participants described that a key factor contributing to the system's effectiveness was this distinction: hard constraints were perceived as non-negotiable requirements, while soft constraints were seen as flexible preferences.

This distinction shaped how participants engaged with the system. Hard constraints provided a foundation for the plan, establishing the core structure, while soft constraints enabled customization and adaptation based on personal needs. As P7 explained: \textit{``So, like, the hard rules are kind of the foundation blocks, then the soft rules are kind of the little building blocks that I can customize in different combinations.''}
Participants found this separation intuitive and essential, reflecting how people naturally distinguish between needs and wants in real-life decision-making. As P4 put it: \textit{``Leveraging those different strictness levels helped manage and streamline the planning of it more, because, you know, we are all busy, right? And in real life, we have a separation between things that you need and things that you want or would be nice to have. I think that is a very useful feature.''}

As shown in Figure~\ref{fig:userStudyQuant2}, different engagement patterns also occurred between \textit{C1 (\ours{})} and \textit{C4 (None)}. For C1, users sorted through their task requirements by going back and forth to the hard and soft constraint bins to organize their constraints. Nine participants defined hard constraints first to ensure that the foundational requirements of the plan were met, then adjusted and adapted specific requirements and preferences through soft rules afterward. They focused on creating an initial outline of the task plan using hard constraints and modifying soft constraints to meet their standards. Five participants used the hard constraints in a reactive manner, transferring preferences in soft constraints to hard constraints after verifying that the preferences were not satisfactorily met in the output plan.

Overall, task completion was much quicker, as shown by the fewer constraint adjustments in \textit{C1 (\ours{})} compared to \textit{C4 (None)} in Figure~\ref{fig:userStudyQuant2}. 
Participants described that this reduction in iterations was largely due to the hard constraints lessening the burden of the user having to verify the outputs against constraint inputs, whereas in C4 every input was effectively a soft preference requiring manual checking.
These results suggest that although incorporating both types of constraints required more initial organization during input, this did not reduce system usability, as shown in Figure~\ref{fig:userStudyQuant}. Instead, the two constraint types supported more effective task completion, resulting in greater usefulness and satisfaction compared to C4.




\subsubsection{Perceptions and Pitfalls of Using Only Hard Constraints}\label{sec:qual2}

\paragraph{Quantitative \& Usage Pattern Results}

The Dunnett's test revealed that \textit{C1 (\ours{})} was significantly more useful $(p=.0074)$ and more satisfying $(p=.0413)$ than the \textit{C3 (Hard Only)} condition. Participants in C1, however, involved significantly more iterations compared to C3 $(p=.0008)$.

Qualitative findings suggest that this dip in usefulness and satisfaction arose from the rigidity and latency of using only hard constraints, coupled with the heightened expectations participants placed on hard rules.

\paragraph{Qualitative Findings}

First, using only hard constraints limited exploration and led participants to a top-down planning style. Participants made fewer attempts to adjust hard constraints in \textit{C3 (Hard Only)} compared to other conditions (Figure~\ref{fig:userStudyQuant2}). Eight participants described spending substantial time upfront to reach an initial version of their plan, but then making a few adjustments. This pattern was attributed to the latency of model checking (approximately three minutes per verification) and the strict, all-or-nothing nature of hard rules. As a result, participants adopted a top-down approach, focusing on finalizing the ``big picture'' requirements rather than iteratively refining details. Several participants noted that because hard rules could not be flexibly modified, they felt less motivated to explore or refine constraints.

Our findings also show that when hard rules worked reliably, participants trusted the system more and expressed stronger satisfaction. Despite the limitations and rigidity of hard constraints, nine participants reported satisfaction with their performance, particularly when they tested the system by toggling specific constraints on and off. They found that the system consistently adhered to their non-negotiable requirements. For example, P3 explained: \textit{``I wanted to see if I could add a formal portrait event in the middle of the wedding. I did one version with the role and one without, and only the version with it consistently checked for it.''} Similarly, in the cooking scenario, P9 highlighted how the system managed complex ordering: \textit{``I input rules like chopping, marinating, and preheating that need to happen before cooking, pretty complex for me to remember on my own. But the plan followed them all. I think it really helps keep things manageable.''} These successful experiences with hard constraints reinforced participants' reliability in the system and their satisfaction with its usefulness.

Finally, qualitative themes show that failures of hard-constraint verification caused disproportionate disappointment and eroded perceived usefulness. Six of the participants reported greater frustration when the system did not behave as expected, either because rules were misinterpreted during the translation stage or because the verifier missed violations in plans that broke multiple hard constraints. Because participants assumed hard rules would be strictly enforced, even a single verification error was judged unacceptable. As P12 noted: \textit{``With the hard violations I would expect a model to try to stick closer to the hard violations than the soft ones. Because a soft rule is like, well, we didn't get this. But it's fine. But missing hard rules will have big consequences.''} Similarly, P4 reflected: \textit{``Hard rules need to be kept no matter what, for the model to not deviate from or be more accurate on. So even though it [hard constraint] was only one rule, it [system] kept missing to check it and that was a deal breaker for me.''}

Overall, while reliably enforced hard constraints bolstered trust and satisfaction, their rigidity and latency discouraged iterative refinement, and any verification failure carried heavier, more consequential weight. Unlike soft constraints, whose occasional misses were expected and more easily tolerated, hard-constraint failures were deemed unacceptable, sharply reducing satisfaction and perceived usefulness.



\subsubsection{Adaptability Support and Explanation Gaps in Soft Constraints}\label{sec:qual3}

\paragraph{Quantitative \& Usage Pattern Results}

The Dunnett's test showed no significant differences between the full system and \textit{C2 (Soft Only)} for perceived usefulness, satisfaction, performance, or usability. In contrast, removing soft constraints in \textit{C3 (Hard Only)} led to a decrease in perceived usefulness and satisfaction, suggesting that soft constraints are critical to the user experience.
The Dunnett's test on constraint-fixing iterations revealed a significant effect, as participants in \textit{C1 (\ours{})} required fewer iterations than in the \textit{C2 (Soft Only)} condition $(p = .0176)$.

Qualitative findings further underscore that soft constraints enabled flexibility crucial for adapting plans to user needs and preferences, while also revealing gaps in explanation quality and verification reliability.

\paragraph{Qualitative Findings}

Interactions with only soft constraints encouraged bottom-up exploration with many iterations. Participants in \textit{C2 (Soft Only)} engaged in numerous interaction attempts, experimenting with different constraint combinations to explore possibilities. Because soft-rule verification responses were quick, users felt free to \textit{P8: ``try things out''} and refine their constraints and plans multiple times. As shown in Figure~\ref{fig:userStudyQuant2}, this resulted in longer task completion times and many more iterations compared to other conditions. While this exploratory process allowed users to filter and select constraints for their final plan, several noted the difficulty of detecting violations and the inconsistent adherence to constraints in outputs forced them to spend additional iterations fixing missed or overlooked rules.

In addition, soft constraints were valued for supporting adaptability and personalization. Seven participants emphasized the value of soft rules in tailoring plans to dynamic needs. P2 described this flexibility as essential for \textit{``aligning vibes''}. Once a baseline plan was generated, iterative feedback allowed participants to gradually shape it to match their preferences. P8, who engaged in the vacation planning scenario, highlighted this adaptability: \textit{``So one day, I wanted my plan to focus on child activities, like waterparks and things like that. But then I felt like the adults maybe wouldn't be getting as much out of it as they would want, so I wanted one day to have more relaxing vibes. But the order of these days doesn't matter and the activities can even be mixed. And the soft rules helped me reflect that in my plan.''} Across scenarios, participants described using soft constraints to customize plans to changing family priorities, moods, party sizes, dietary needs, themes, or situational factors like weather. They appreciated the ability to guide and adjust the system, describing this flexibility improved the alignment of outputs with their expectations.

On the other hand, qualitative themes show that soft-only explanations lacked clarity and reliability, limiting their influence on decision-making. Five participants reported that explanations of soft constraints were often verbose, jargon-like, or ``opinion-like'' rather than actionable. As P9 explained: \textit{``It isn't an explanation, but more like their thought on telling me how well-aligned it seems to be. I don't need someone to tell me how to think. I know what my preferences are, and I can get what they are giving as an explanation by just reading the plan itself.''} Five other participants also noted that these explanations had little impact on their decision to select a plan, and some questioned the reliability of the rating system. As P2 commented: \textit{``Between giving me five stars and saying that it's a great plan, it also broke some hard rules. So how could it be five stars if it didn't do everything it was supposed to do? Does the system not know that?''} These inconsistencies created an additional burden for users to verify adherence themselves, reinforcing the perception that soft-only explanations could support early ideation but were insufficient for final decision-making.

Overall, while soft constraints were valuable for adaptability and personalization, weak explanations and verification gaps undermined their reliability and usefulness alone. This tradeoff indicates that users valued soft constraints for flexibility but did not fully trust them to ensure correctness, limiting their role in supporting user confidence when selecting a final plan.

\begin{table*}[!b]
\centering
\caption{Participant ID (PID), occupations, domains, years of experience, and selected planning tasks of expert participants (Study~2).}
\label{tab:experts}
\small
\renewcommand{\arraystretch}{1.3} 
\begin{tabular}{@{}l p{3.8cm} p{2.5cm} c p{6.2cm}@{}}
\toprule
\textbf{PID} & \textbf{Occupation} & \textbf{Domain} & \textbf{Years} & \textbf{Planning Task} \\
\midrule
E1 & Graduate Program Manager & Higher Education & 9 & Planning graduate student welcome week \\ \addlinespace
E2 & Undergraduate Program Manager & Higher Education & 3 & Organizing undergraduate graduation and commencement \\ \addlinespace
E3 & Academic Advisor & Higher Education & 10 & Coordinating student organization fair \\ \addlinespace
E4 & Grant Post-Award Accountant & Finance & 5 & Projecting and allocating grant budgets \\ \addlinespace
E5 & Senior Superintendent (Construction Company) & Construction Management & 32 & Scheduling construction phases, subcontractor tasks, and equipment use on a project site \\ \addlinespace
E6 & Interior Designer & Design / Architecture & 4 & Scheduling construction tasks and coordinating collaborators for bathroom renovation project \\
\bottomrule
\end{tabular}
\end{table*}

\section{User Study 2: Domain Experts}\label{sec:userStudy2}

To assess \ours{} in professional contexts where planning is more complex and high-stakes, we conducted a second study with domain experts to examine its utility, reliability, and alignment with expert workflows.

\subsection{Method}

\subsubsection{Study Design}
This study examined how well \tool{} supports real-world planning tasks in professional contexts. We recruited experts from diverse domains where planning tasks play a central role in their professional workflows. 
Each session began with a short interview to learn about the expert's domain and typical planning tasks. Based on this discussion, the expert selected a planning task to carry out with \tool{}, described in Table~\ref{tab:experts}. They then freely engaged with the system, entering hard and soft constraints as well as prompts relevant to their task. After completing the interaction, experts filled out quantitative scales and participated in semi-structured interviews. Each session lasted approximately one hour.

\subsubsection{Participants}
A total of six participants were recruited for the expert user study. Eligibility criteria included being located in the United States, fluency in English, and a minimum age of 18. All participants were recruited through university mailing lists and snowball sampling.

Participants age ranged from 22--57 ($M = 40.3$, $SD = 12.5$). 66.6\% of the participants identified as female and 33.4\% as male. 83.3\% of our participants identified as White and 16.7\% as Asian. Table~\ref{tab:experts} lists participants' occupations, domains, and years of experience. After the study, participants were compensated \$15.00 per hour. We refer to participants as E1--E6, using the notation E\textit{i} to indicate expert participants, where \textit{i} indicates participant ID number.

\subsubsection{Measures \& Analysis}
To assess experts' perceived performance, usefulness, satisfaction, and usability of the system, we employed the same measures used in the first user study described \S\ref{sec:userStudy_method}. 

\begin{figure}[!tb]
  \includegraphics[width=4in]{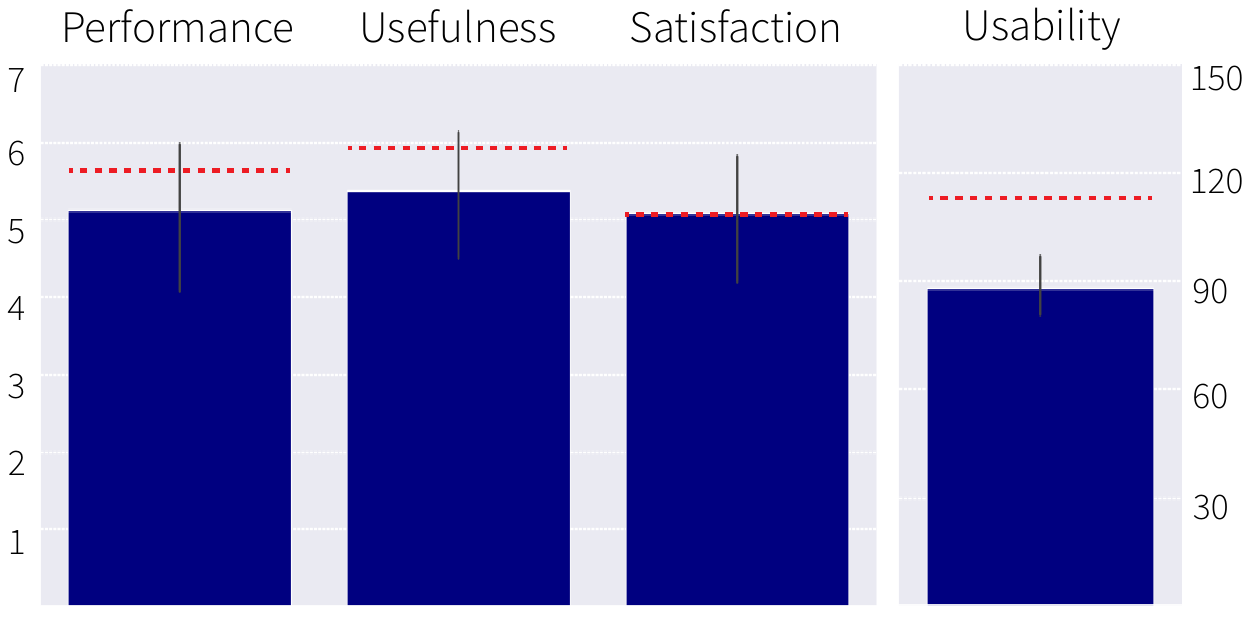}
   \vspace{-12pt}
  \caption{\textit{Quantitative Data from User Study 2 ---} Bar graphs on experts' perceived performance, usefulness, satisfaction, and usability scores across different conditions. Red lines represent results from Study 1, overlaid for context. Vertical lines in each bar graph indicate standard error. 
}
  \label{fig:userStudyQuant3}
\end{figure}

For quantitative analysis, we visualize the results of the experts in Figure~\ref{fig:userStudyQuant3} with \textit{C1 (\ours{})} results from Study~1 on the same graphs to contextualize the experts' outcomes. For qualitative analysis, we conducted a thematic analysis of interview data, following the same procedure described in \S\ref{sec:userStudy_method}.

\subsection{Findings}

Our analysis aimed to understand how domain experts engaged with \ours{} in their professional planning tasks, with attention to its usefulness, usability, and fit within expert workflows. Based on quantitative, qualitative, and objective measures, we present three main findings: (1) \ours{} addressed experts' struggles with LLMs by aligning with their need to manage strict rules; (2) experts envision cooperative worker–AI workflows with \ours{}, reducing undesirable and time-consuming tasks; and (3) experts desired support for reducing context overhead and streamlining constraint input.

\subsubsection{\ours{} Addressed Experts' Struggles With LLMs by Aligning With Their Need to Manage Strict Rules}

Five experts found the system useful because it alleviated many burdens they had experienced with traditional LLMs. Four experts described prior difficulties: task requirements were hard to fully capture in a prompt, verifying outputs against constraints was burdensome, generating satisfactory plans required extensive trial and error, and outputs often contained hallucinations or constraint violations that reduced usability. They emphasized that \tool{} addressed these challenges by enabling them to define, manage, and enforce rules tailored to their professional planning tasks.

Within \tool{}, experts especially valued the distinction between hard and soft constraints, though they described how they relied far more on hard constraints for professional tasks. They found it easier to generate hard constraints quickly, while soft constraints were harder to articulate and often left blank. Hard constraints matched the professional character of their tasks, covering regulations, budgets, deadlines, procedural rules, and resource limits. These were seen as straightforward, critical requirements for plan success, whereas soft constraints introduced unnecessary subjectivity and were considered less essential. Still, three experts noted soft rules were helpful for handling uncertainty or minor flexibility within boundaries of hard constraints (\eg table settings for different audiences).

Four experts' workflow with \ours{} began by entering hard rules, which helped them mentally organize requirements before crafting a planning prompt. The other two experts started by drafting their prompt and hard constraints simultaneously. When reviewing plans, experts focused mainly on hard-rule explanations. Participants found the listing of violated hard constraints to be a quick, factual validation of whether a plan could succeed, enabling actionable next steps. In contrast, soft explanations were often dismissed as \textit{E2: ``providing uninvited opinions''} that did not add to reliability. Two experts even suggested reordering the interface to present hard-rule violations before the generated plan, allowing them to run a quick ``sanity check'' before deciding whether to review the rest of the plan.

Finally, experts' hard rules often extended beyond the system's LTL-based design. While some constraints fit into LTL properties (\eg ``paint the walls before installing the toilet'' or ``no work may begin before permits are approved''), many were either simpler or more complex---a pattern more pronounced than in Study~1. Simpler constraints included numeric limits (``stay within a \$3500 budget''), cardinality requirements (``at least one faculty speaker''), and categorical inclusions (``the event must include food and beverage''), which require arithmetic or relational checks rather than temporal reasoning. More complex rules, such as duration-based deadlines (``permits take one month to be approved'') or calendar-specific conditions (``the event must be held during commencement week''), align better with formalisms like Simple Temporal Networks (STNs) or Metric Temporal Logic (MTL). This mismatch between the types of constraints experts required and those the system could verify led to verification failures more often than in Study 1. These failures likely contributed to experts rating the system's performance lower than general users, as several of their constraints could not be fully verified within LTL. Overall, these patterns suggest that expert-specified hard rules span a broader range of property types, highlighting the need for expanded verification approaches.

\subsubsection{Experts Envision Cooperative Worker–AI Workflows With \ours{}, Reducing Undesirable and Time-Consuming Tasks}

Participants highlighted the \ours{}'s value in fostering collaboration between themselves and the LLM. Five participants noted that the generated plans often contained useful ideas they had not considered, making the system helpful for brainstorming and especially valuable when planning for unfamiliar requirements or new projects. Three participants emphasized that the system also eased the challenge of adapting plans as constraints evolved. They explained that, in practice, new rules or unexpected circumstances frequently arise, but with the system they could quickly adjust inputs and regenerate plans (or reuse earlier versions as prompts) to produce updated solutions.

In both cases, participants emphasized that the system reduced the burden of tedious but important tasks such as generating ideas, organizing plans, and managing constraints. These activities were often described as undesirable and time-consuming, and the system's ability to provide and verify draft plans gave workers a foundation to build on, reducing the sense of overwhelm. One participant described, \textit{E3: "It helps me get over the initial chaos---the mess of constraints, new requirements, and who or what is involved. Once that's cleared, I can sit down and think, okay, this might work, this doesn't. Clearing that mess lets me see where I actually need to spend my time to make the plan work or improve it. Without that first step, I just get stuck in the mess."}

Looking forward, all six experts envisioned cooperative workflows where the system plays an assistive or supporting role. They imagined the LLM assisting with idea generation, constraint management, plan structuring, and verification, while workers retained the role of refining, overriding, and adapting the output. In this vision, experts similarly described that the LLM would take the first pass at organizing messy constraints and providing options, offering high-level guidelines that workers could then shape to fit their professional needs.

\subsubsection{Experts Desired Support for Reducing Context Overhead and Streamlining Constraint Input}

Unlike general users in Study 1, all six experts consistently struggled with the difficulty of collecting and entering the large number of hard constraints their planning tasks required, as well as conveying background domain knowledge the system lacked.

First, participants noted that remembering and inputting numerous hard constraints created context overhead. Experts noted that each planning task required many rules, but these often operated on ``autopilot'' because they had carried them out so many times, or because the high-level structure was similar across tasks. As a result, experts struggled to recall every hard constraint offhand, and the volume of rules was too large to input manually. This need to remember and enter constraints was described as a source of significant context overhead.

In addition, participants highlighted that domain knowledge gaps in \ours{} further added to the overhead. Four participants emphasized that they did not know how much background knowledge had to be supplied for the system to plan effectively. They assumed the system would share common professional context, but it often did not. For example, one expert (E2) noted that the system interpreted ``one year'' as twelve months instead of the nine-month ``academic year,'' while another (E6) was surprised when the system scheduled a full week for toilet installation, which is typically a one-day task. In response to these gaps and study time, four experts simplified their planning tasks and constraints, and described this contextual overhead as the main factor reducing usability, usefulness, and satisfaction as shown in Figure~\ref{fig:userStudyQuant3}.

To address these challenges, participants proposed ways to streamline constraint input. Five participants suggested template- or protocol-based approaches, where frequently used or repetitive constraints could be filled in semi-automatically. Other three participants envisioned a learning process, where the system would save plans and constraints, gather feedback from the user on the plan outcome, and reuse them as protocols for future interactions. Two others suggested integrating \ours{} with external tools such as Excel to streamline the constraint input process by automatically importing constraints for users to review and edit.

\section{Discussion}
Our findings from two user studies highlight the central challenge of combining reliability and flexibility in LLM-based planning. General users and domain experts alike recognized the value of balancing strict enforcement of critical rules with the adaptability needed to accommodate preferences and evolving contexts. Study 1 showed that explicitly distinguishing hard from soft constraints increased perceived performance, usefulness, and satisfaction, and crucially, this added value did not come at the expense of usability. Hard rules improved perceived trustworthiness, while soft rules supported personalization, with the combination seen as both effective and manageable. At the same time, Study 2 revealed that experts often relied on extensive sets of hard constraints (many extending beyond LTL's expressiveness) and faced new burdens in recalling and specifying all relevant rules. These results suggest that the design of constraint-based interaction with LLMs must go beyond binary enforcement to support flexible strictness, richer coverage of constraint types, and interfaces that ease the overhead of constraint input.

Together, the two studies underscore a broader framing: effective user–AI planning systems must give users, not the system, the authority to define their own constraints and distinguish between types. This requires not only technical verification methods, but also interaction designs that help users articulate, manage, and refine their constraints without excessive burden. Additionally, our results highlight the importance of actionable explanations when rules are violated, hybrid verification approaches that extend beyond LTL, and features that help users capture and reuse domain knowledge. We next outline design implications that address these challenges and opportunities for supporting constraint flexibility in practice.

\subsection{Support User Control by Allowing Both Hard and Soft Constraints}

Both studies confirmed that allowing users to mix hard and soft constraints yields better outcomes. General users found that hard rules provide a non-negotiable foundation while soft rules act as adaptable ``building blocks'' to tailor plans to personal needs. This flexibility led to significantly higher perceived performance, usefulness, and satisfaction compared to having no constraints. The pattern aligns with the soft-constraints literature and preference modeling, where graded or conditional preferences complement strict requirements to handle real-world variability \cite{boutilier2004cp, bistarelli1997semiring}. More recently, \citet{liu2024we} observed that practitioners often need ``low-level'' constraints (\eg specific formats or timing that must be followed) alongside ``high-level'' preferences (\eg softer guidelines on style or content) to effectively integrate AI into developer workflows. In practice, supporting varying constraint strictness (from strict rules to looser preferences) enables LLM planners to accommodate real-world complexity and shifting user needs, improving trust and effectiveness.

\paragraph{\textbf{Design Implication:} Support users to define and balance both hard and soft constraints. }Our findings suggest that systems should give users direct control and tools (\eg categorization, sliders, toggles, or checklists) to explicitly declare which rules are non-negotiable hard constraints and which are softer preferences. Supporting a range of constraint types can enable users to structure planning tasks in ways that best fit their goals and environments, leading to outputs that are better aligned with personal needs. Prior research on AI steering tools similarly found that when people, including novices, can adjust a generative AI's parameters or rules, they achieve results that feel more fitting and creative \cite{lee2025veriplan}. Therefore, future LLM-based planning systems should include interaction mechanisms that let users define, mix, and prioritize constraints.


\subsection{Design for Constraint Hierarchies and Task-Adaptivity}
\revision{
Our findings highlight that while enabling both hard and soft constraints offers users an effective means to express priorities, configurations with only one type reveal important design trade-offs. When users could only use hard constraints, their experience suffered from rigidity and slower iteration, whereas soft-only configurations lacked sufficient reliability. These findings suggest that future systems should not treat constraint types as binary categories but as a spectrum of prioritization and enforceability.

\paragraph{\textbf{Design Implication:} Enable adaptive and hierarchical constraint specification} 

From a design perspective, allowing users to flexibly reclassify, rank, or layer constraints (\eg specifying high-, medium-, or low-priority rules) may support more natural expression of importance and better adaptation across contexts. On the other hand, systems that must rely on a single constraint type should adapt mechanisms to preserve the missing qualities of the other (\eg speeding up verification and improving interpretability for hard-constraint–only settings, or adding lightweight verification or symbolic consistency checks to strengthen reliability for soft-constraint–only systems). More broadly, future research could explore how constraint-based interaction can evolve beyond the soft–hard dichotomy toward adaptive constraint hierarchies, where the system learns to adjust enforcement levels dynamically based on task stakes, user behavior, or prior outcomes. Such approaches would move constraint-based user–AI interaction from static specification toward fluid prioritization and adaptive enforcement, better reflecting how people naturally balance flexibility and reliability in real-world decision-making.
}
\subsection{Use Hard Constraints Sparingly Given Higher Expectations}

Our findings revealed that hard constraints shaped different user expectations compared to soft constraints. Because users saw hard constraints as foundational, non-negotiable requirements, they assigned greater importance to them. Consequently, violations of hard constraints were perceived as more severe, leading to a sharper decline in satisfaction, perceived performance, and overall system usefulness.

\paragraph{\textbf{Design Implication:} Hard constraints must be used sparingly and carefully.} These results suggest that hard constraints should be implemented sparingly and deliberately. When systems apply hard constraints too broadly and especially if they are frequently violated, users may experience broken expectations, which in turn diminishes trust and satisfaction. This pattern aligns with prior work showing that expectation violations (when system outcomes fail to match what users believe was promised or implied) significantly reduce trust \cite{kizilcec2016much}. Therefore, hard constraints should be reserved for situations where dependability is critical and should be reliably upheld. 

Given that some violations may still occur, systems should also include mechanisms to mitigate their impact. Providing explanations for why a rule was broken, listing which constraints were violated, or offering guidance on how users can modify their constraints or requests can help preserve users' perception of the reliability and usefulness of the system.

\subsection{Explanations Should Map to Violated Rules, Not Generic Ratings}

Users in both studies emphasized the importance of explanations that clearly map to the rules they defined. Hard constraint explanations that explicitly listed violated rules were valued more highly and deemed more useful, as they were straightforward, intuitive, and made the violations actionable. In contrast, the star ratings and verbose reasoning used for soft constraints were often seen as confusing, misaligned with user judgment, or unnecessary. While users acknowledged that it gave a general sense of the plan's adherence to constraints, they also noted that the rating often failed to align with their own judgment, especially when a plan with multiple hard-constraint violations still received a high rating. This mismatch made users question the validity of the system's explanation, consistent with HCI findings that generic transparency or scalar scores often fail to improve human–AI outcomes \cite{poursabzi2021manipulating}.

\paragraph{\textbf{Design Implication:} Provide actionable explanations that avoid simplistic quality assessment. }Our findings suggest that more advanced methods (\eg analysis of alignment to user preferences and plan repair cost) are needed to make quality judgments about plans, and, in the absence of such methods, systems should not utilize simplistic evaluations. Users are capable of evaluating plans themselves and often have values or preferences that the system cannot fully consider. Future systems should avoid simplistic scoring and instead provide explanation interfaces that highlight exactly which user-specified constraints were satisfied or broken, along with concrete suggestions for how users might refine their rules or adjust their plans. For instance, future interfaces can list violated hard rules and provide suggestions on how to revise the rules or modify their constraints to address the issue, and previews of how these changes may affect the planning outcome. Providing such actionable guidance can help users refine their plans and get closer to their desired outcomes.
\subsection{Assist Users with Constraint Input to Reduce Overhead}

Our second user study revealed a new challenge where experts struggled with the ``context overhead'' of supplying the many constraints and domain conventions their tasks required. Planning often involved dozens of hard rules and implicit practices (\eg standard timelines, common procedures), but recalling and entering each detail was tedious and error-prone. Compounding this, the LLM sometimes misinterpreted domain concepts (\eg treating an academic year as 12 months instead of nine, or overestimating task durations), which forced users to simplify their inputs and reduced the system's usefulness. This aligns with HCI evidence that free-form prompting imposes cognitive load and leads to mismatched expectations \cite{zamfirescu2023johnny, subramonyam2024bridging}. To address these issues, future systems should actively support constraint specification. Participants suggested templates or structured forms for common rules, storing user-defined constraints as reusable ``profiles,'' and integrating with external tools (\eg spreadsheets, project management apps) to import known constraints automatically. Such smarter authoring interfaces and memory can reduce context overhead, ensuring usability while preserving critical domain knowledge.

\paragraph{\textbf{Design Implication:} Assist users with constraint input and context management to reduce overhead}

Our findings suggest that future systems to build constraint authoring around reuse, structured capture, and import. This authoring support can be provided through domain-ready templates and protocols for recurring rules, allowing users to save finalized rule sets and outcomes as reusable ``profiles'' for future projects and offering importers for spreadsheets, calendars, and project tools to pre-fill known parameters for review. Additionally, systems can add guidance for hidden domain assumptions by prompting clarifications on ambiguous terms (\eg ``Is an academic year 9 months?'') and by highlighting suspicious durations, which experts struggled to correct post-hoc. Finally, systems can combine this support and guidance into cooperative workflows, where the system drafts an initial, organized set of constraints and the user refines or overrides them, reducing context overhead without ceding control. These cooperative workflows embody HAI guidelines to remember past interactions, learn from user behavior, and make reasons clear \cite{amershi2019guidelines}.
\subsection{Integrate LLMs to Broaden Verification While Preserving Usability}

\ours{} shows that LLMs can automate the translation from natural language into model checking properties, such as LTL rules and PRISM code. Unlike template-based approaches \cite[\eg][]{lee2025veriplan}, LLMs expand coverage by handling a broader range of user-defined rules, including those not easily captured in formal logic. This enables users to freely express hard constraints in natural language while still benefiting from formal verification.

This flexibility is especially valuable for domain experts, whose rules often require numeric, role-based, or contextual reasoning beyond simple temporal order. By augmenting user-declared rules with LLM translation, systems can extend verification power while keeping users in control of constraint types and priorities. This approach aligns with recent findings that LLMs are poor self-verifiers, and thus should be coupled with external, sound verifiers to ensure reliability \cite{kambhampati2024llms, stechly2024self}.

Importantly, our study suggests that this added capability did not come at the expense of usability. Our user study suggests that users did not perceive increased difficulty or reduced usability when engaging with these system designs, indicating that integrating LLM translation preserves ease of use while making verification more accessible. Here, accessibility means not only lowering barriers to entry (\ie removing the need to learn formal specification languages) but also broadening what users can accomplish, narrowing the gap between expert verification tools and everyday end-user planning needs.

\paragraph{\textbf{Design Implication:} LLM integration in model checking can improve generalizability and accessibility. }

As \ours{} demonstrates, future systems can use LLMs as translators, to provide an assistive layer atop formal tools, and support verifiable round trips via model checking to enable formal verification in LLM-based tasks. However, users' usability and usefulness are important to successfully implement model checking. To do so, designers should consider: (1) show the mapping from user phrases to formal predicates and back-translate finalized properties into plain language for user confirmation before verification. Prior work on end-user programming similarly indicates that providing formal feedback in a user-friendly manner can improve outcomes. For example, \citet{porfirio2024goal} found that a visual ``plan checker'' component (which validated user-authored robot plans against goals) significantly improved plan quality without overwhelming users; (2) Surface a verification-coverage indicator when a rule can't be fully expressed in LTL (\eg numeric, relational, or contextual clauses): formally verify the LTL-expressible portion and label the remainder as heuristic checks so users know exactly what wasn't proved; and (3) preserve usability with lightweight controls interface features, by showing the original text, the LTL/PRISM form, and the translations side-by-side, along with an intuitive, easy ``revise'' loop. Such patterns can widen the set of constraints users can successfully bring into verification while keeping the interaction approachable for both general users and experts.

\section{Limitations \& Future Work}
Our work has several limitations. First, our chosen model checking framework, particularly PRISM and Stormpy, limits the expressiveness of temporal constraints or logical expressions.
Future work can explore more expressive options for off-the-shelf model checkers or verification techniques, such as those that accept computation tree logic (CTL) constraints.
In addition, the evaluation datasets used in our component-level evaluation are modest due to needing to create them by hand. Future work can focus on further investigating the applicability of LLMs in model-checking procedures on full datasets or a wider range of scenarios. Similarly, our empirical evaluation is limited to data from 18 participants. Future work can explore larger-scale evaluations to more conclusively validate the outcomes presented in this paper and build on the results to highlight additional insights. 
Finally, future work can involve the expansion of \ours{}. For example, the system can be modified to enable users to ablate or modify the five-star rating system. Another direction can involve incorporating different machine learning techniques, such as retrieval-augmented generation (RAG) \cite{lewis2020retrieval}, to help users retrieve and build upon previous plans.

\section{Conclusion}

We introduce \ours{}, a system that enables users to explicitly define and verify both hard and soft constraints in LLM-based planning tasks. A central contribution of \ours{} is placing control in the hands of users: they, not the system, decide which rules must be enforced as hard constraints and which can remain flexible as soft preferences. By supporting this distinction, \ours{} provides reliability where guarantees are needed while preserving adaptability where flexibility is valuable.

Through two user studies with general users and domain experts, we found that supporting user-defined constraint types improved perceived performance, usefulness, and satisfaction. Hard and soft constraints were seen as serving distinct but complementary roles, with violations of hard constraints undermining heightened expectations. At the same time, experts surfaced challenges in managing extensive sets of rules and domain-specific knowledge, underscoring the need for interfaces that reduce input overhead and broaden constraint coverage beyond LTL.

Together, these findings demonstrate that effective planning support requires systems to give users control over how constraints are expressed and enforced. We present a system that operationalizes user-defined constraint types, techniques that enable formal verification through integration with LLMs, and design implications for how future systems can strengthen reliability and better align outputs with user intent. By centering user control in constraint specification, this work advances toward planning systems that are both dependable and adaptable to the complexities of real-world use.
\bibliographystyle{ACM-Reference-Format}
\bibliography{bibliography}

\end{document}